\documentclass[preprint,12pt]{elsarticle}

\usepackage{graphicx,subcaption}%
\usepackage{multirow}%
\usepackage{amsmath,amssymb,amsfonts}%
\usepackage{amsthm}%
\usepackage{mathrsfs}%
\usepackage[title]{appendix}%
\usepackage{xcolor}%
\usepackage{textcomp}%
\usepackage{manyfoot}%
\usepackage{booktabs}%
\usepackage{algorithm}%
\usepackage{algorithmicx}%
\usepackage{algpseudocode}%
\usepackage{listings}%
\usepackage{amsmath}
\usepackage{pifont}
\usepackage{rotating}
\usepackage{caption}
\usepackage{placeins}
\usepackage{makecell}
\usepackage{multicol}
\usepackage{hyperref}
\usepackage{placeins}
\usepackage{longtable}
\usepackage{lipsum}
\usepackage{tabularray}
\usepackage{tabularx}
\newcommand{\xmark}{\ding{55}}%
\newcommand{\tabitem}{~~\llap{\textbullet}~~}
\usepackage{adjustbox}
\usepackage{microtype}

\usepackage{xspace}
\def\XS{\xspace}
\DeclareMathAlphabet{\mathb}{OML}{cmm}{b}{it}
\def\sbm#1{\ensuremath{\mathb{#1}}}                % Style gras italique (necessite amsmath)           
\def\scu#1{\ensuremath{\mathcal{#1\XS}}}   

\newcommand{\etal}{\textit{et al. }}
\newcommand{\etc}{\textit{etc. }}
\def\cf{\textit{cf.}\XS}
\def\eg{\textit{e.g.,}\XS}

\def\Eb{{\sbm{E}}\XS}
\def\Xc{{\scu{X}}\XS} 
  
\def\zb{{\sbm{z}}\XS}
\def\figureabvr{Fig.\XS} 
\def\tableabvr{Table\XS}
\usepackage{tablefootnote}
\usepackage{tablefootnote}
\usepackage[para]{footmisc}
\usepackage{xcolor, soul}
\sethlcolor{yellow}
\usepackage{color}
\def\rev#1{{\color{blue}{#1}}}

\journal{Artificial Intelligence in Medicine}
\begin{document}

\begin{frontmatter}

\title{Deep Generative Models for Physiological Signals: A Systematic Literature Review}

\author[redcad]{Nour~Neifar\corref{mycorrespondingauthor}}
\ead{nour.neifar@redcad.org}
\author[redcad]{Afef~Mdhaffar} 
\ead{afef.mdhaffar@enis.tn}
\author[crns]{Achraf~Ben-Hamadou}\ead{achraf.benhamadou@crns.rnrt.tn} 
\author[redcad]{Mohamed~Jmaiel} 
\ead{mohamed.jmaiel@redcad.org}

\address[redcad]{ReDCAD Lab, ENIS, University of Sfax, Tunisia}
\address[crns]{Centre de Recherche en Num\'{e}rique de Sfax, Laboratory of Signals, Systems, Artificial Intelligence and Networks, Technopôle de Sfax, Sfax, Tunisia}
\cortext[mycorrespondingauthor]{Corresponding author}

\begin{abstract}
In this paper, we present a systematic literature review on deep generative models for physiological signals, particularly electrocardiogram (ECG), electroencephalogram (EEG), photoplethysmogram (PPG) and electromyogram (EMG). Compared to the existing review papers, we present the first review that summarizes the recent state-of-the-art deep generative models. By analyzing the state-of-the-art research related to deep generative models along with their main applications and challenges, this review contributes to the overall understanding of these models applied to physiological signals. Additionally, by highlighting the employed evaluation protocol and the most used physiological databases, this review facilitates the assessment and benchmarking of deep generative models. 
\end{abstract}

%%Graphical abstract
%\begin{graphicalabstract}
%\includegraphics{grabs}
%\end{graphicalabstract}

%%Research highlights
% \begin{highlights}
% \item Research highlight 1
% \item Research highlight 2
% \end{highlights}

\begin{keyword}
%% keywords here, in the form: keyword \sep keyword

Deep generative models, ECG, EEG, PPG, EMG

\end{keyword}

\end{frontmatter}

%% \linenumbers

\section{Introduction}\label{sec:sec1}
Physiological signals (\eg ECG, EEG, PPG, EMG) and medical imaging (\eg MRI, CT) present an essential tool in health monitoring as they provide significant information about the body's internal state \citep{lanza2007electrocardiogram,muhammad2020eeg}, \citep{kloppel2012diagnostic,risacher2021neuroimaging}. 
Recently, deep learning methods have attracted significant interest to analyze physiological signals \rev{and medical imaging} for diagnosis and treatment purposes \citep{zhao2019deep,zhang2020ecg,miao2020continuous,miften2021new},\rev{\citep{jo2019deep,alsubaie2024alzheimer}}.

In particular, deep generative models have gained significant attention and have been effectively used in the medical field for various tasks \citep{ISLAM2023109201,chen2018deep,hwang2023real}. 
\rev{Despite the existence of extensive reviews that analyze the use of generative models for medical imaging \mbox{\citep{wang2023applications,laino2022generative,huynh2024review,kazerouni2023diffusion,zhao2025diffusion,ali2022role}}, fewer studies systematically explore their use for medical time series data like physiological signals, leaving a significant gap in the literature. Therefore, it is crucial to provide a systematic literature review that focuses on analyzing research studies dealing with generative models applied to physiological signals; which is the main scope of this paper. Indeed, the existing reviews on physiological signals \mbox{\citep{monachino2023deep,berger2023generative}} primarily focus on ECG signals and do not cover a large spectrum of generative models. For example, Berger \etal \citep{berger2023generative} specifically focuses only on Generative Adversarial Networks (GANs)-based approaches, further limiting the scope of this review. Other reviews have explored deep generative models applied to time series data.} For instance, Brophy \etal \cite{brophy2023generative} provided a comprehensive overview of GANs, specifically for the analysis of time series data. The main objective of their paper is to summarize the current discrete and continuous variants of GANs as well as their challenges in the context of time series. 
Zhang \etal \cite{zhang2022comprehensive} conducted a comprehensive review on GANs applied to time series such as speech, music, and biological signals. They summarized the latest advancements for the generation of these signals using GANs, as well as the existing assessment protocols employed to evaluate the GANs' performance. %Musa \etal \cite{musa2022systematic} presented a systematic review on the use of deep learning algorithms in ECG analysis. They have also performed a meta-data analysis that summarizes the areas, advantages, and challenges of applying deep learning models to ECGs. 
Given the importance of physiological signals in human health monitoring, it is valuable to explore the challenges and opportunities they present, especially in relation to the application of deep generative models. To the best of our knowledge, this paper presents the first systematic literature review on the application of deep generative models with a focus on the essential and commonly used physiological signals, in particular the electrocardiogram (ECG), electroencephalogram (EEG), photoplethysmogram (PPG), and electromyogram (EMG).

\begin{figure*}
    \centering
    \includegraphics[width=\textwidth]{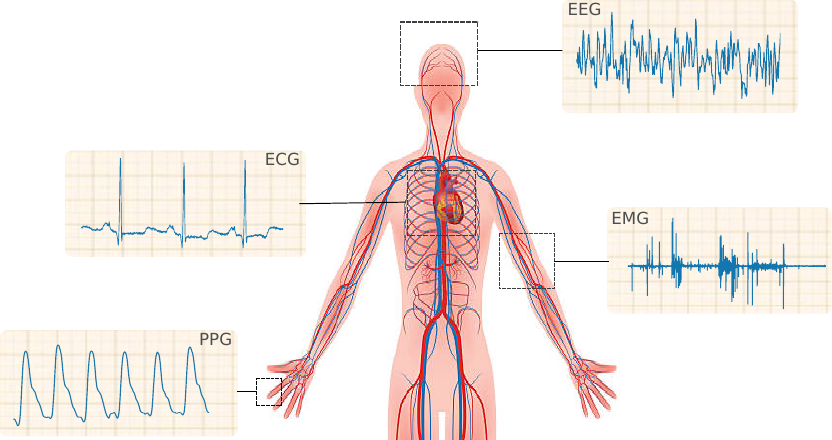}
    \caption{Samples of physiological signals from clinical datasets: \rev{ECG, EEG, PPG, and EMG.}}
    \label{fig:signals}
\end{figure*}

ECG signals represent the electrical activity of the heart, captured by electrodes placed on the chest and limbs. ECGs are often employed for the diagnosis of heart disorders and monitoring of heart activity. On the other hand, EEG signals represent the electrical activity of the brain recorded by placing electrodes on the scalp. EEGs provide important insight related to brain activity, including neurological disorders. PPG signals measure the hemodynamic activity of the heart. These signals, which are often measured at the body periphery such as the fingertip, offer important information about the cardiovascular system. While EMG signals are recorded by using electrodes positioned on the muscles to measure their electrical activity. These signals represent a valuable tool for diagnosing neuromuscular disorders as well as assessing muscle function and activity during different tasks or movements. \figureabvr \ref{fig:signals} represents examples of ECG, EEG, PPG, and EMG signals taken from real databases. 

Our objective is to present a comprehensive overview of the current state-of-the-art deep generative models currently used in the analysis of the discussed signals. By conducting a thorough analysis and synthesis of the existing literature guided by well-defined research questions, we aim to identify the various deep generative architectures employed in analyzing physiological signals. We explore how these models have been applied to address problems with physiological signals. Furthermore, we identify and discuss the challenges faced by deep generative models in analyzing physiological signals. Additionally, we review the existing evaluation protocols and metrics used in the literature to assess the performance of deep generative models on the most widely used physiological databases in this field. This synthesis can help researchers to select appropriate models, address challenges, and explore future directions for advancing the field.

The rest of this paper is structured as follows. In Section \ref{sec:sec2}, we outline the adopted methodology for conducting our systematic literature review (SLR). We describe the search strategy, as well as the inclusion and exclusion criteria, and the data extraction process. Section \ref{sec:sec3} presents the results of our SLR and provides an analysis of the identified studies. In section \ref{sec:discussion}, we discuss the findings and some direction for future research. In section \ref{sec:sec4}, we present a summary of our paper.

\section{Methodology}\label{sec:sec2}
In our systematic literature review, we followed a well-defined methodology that included the following elements:
\begin{enumerate}
    \item Formulation of specific research questions to address the aims of our study,
    \item Development of a comprehensive search strategy to identify relevant research,
    \item Definition of inclusion and exclusion criteria to select the studies that could be considered in our review,
    \item Collection of data. 
\end{enumerate}

\subsection{Research questions}
In our systematic literature review, we consider the following research questions (RQs).
  
\begin{itemize}
\item RQ1: What are the most commonly used classes of deep generative models for ECG, PPG, EEG, and EMG signals?
\item RQ2: How are these classes of deep generative models applied in practice?
\item RQ3: What are the main challenges associated with using deep generative models for ECG, PPG, EEG, and EMG signals?
\item RQ4: What is the commonly used evaluation protocol for assessing the performance of deep generative models?
\item RQ5: Which physiological datasets have been utilized to evaluate the effectiveness of deep generative models?
\end{itemize}

\begin{figure*}[t]
\centering
\includegraphics[width=0.9\textwidth]{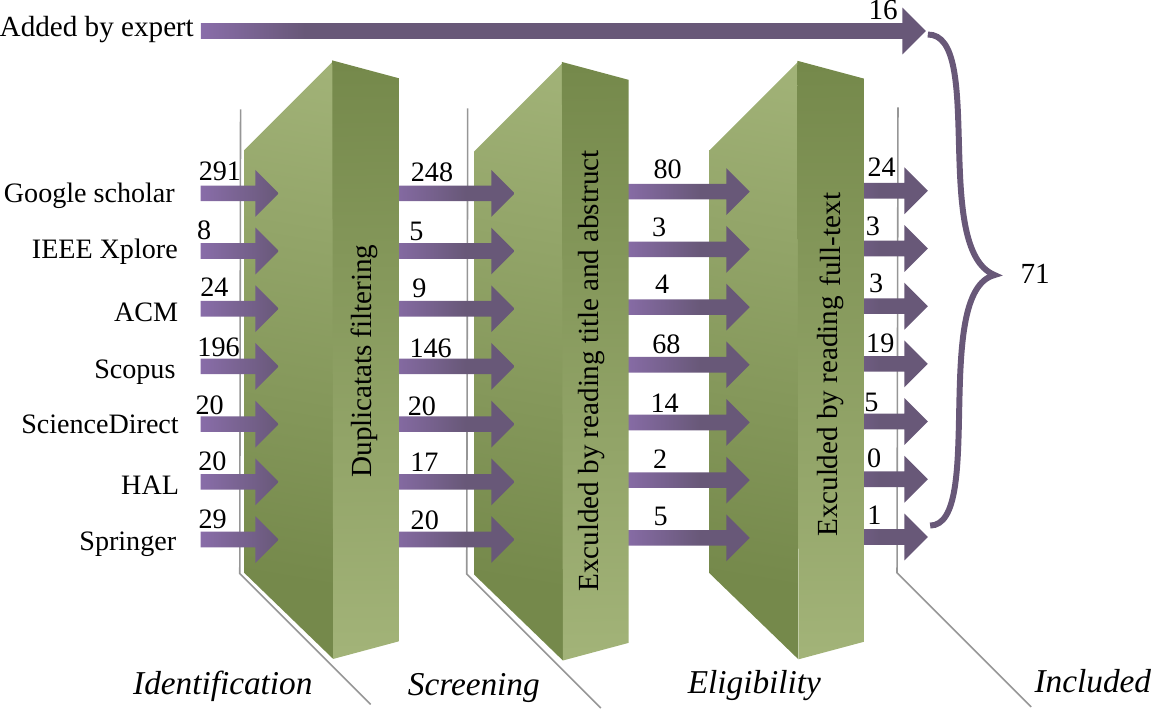}
\caption{Search methodology used for paper selection in our systematic review. From a total of 588 studies, we selected 55 studies through three exclusion steps, and we also included an additional 16 studies recommended by the expert.}
\label{fig:data_collection}
\end{figure*}
\subsection{Query}
To capture relevant literature for our systematic review, a search was conducted between 2018 and 2023 in various search engines, including Google Scholar, IEEE Xplore, ACM Digital Library, Scopus, ScienceDirect, HAL and Springer. We first defined a set of keywords based on the research questions. Next, these keywords were combined using boolean operators such as AND and OR to formulate the following search query:

(``electrocardiogram" AND ``deep generative models") OR (``electroencephalogram" AND ``deep generative models") OR (``photoplethysmogram" AND ``deep generative models") OR (``electromyogram" AND "deep generative models")

\subsection{Inclusion and exclusion criteria}
We established different criteria to ensure that the selected papers align with the research questions and objectives of our systematic review.
\begin{itemize}
    \item Papers that correspond to a search term are considered,
    \item Only the signals modalities of ECG, EEG, PPG, and EMG are considered,
    \item Papers published between 2018 and 2023 are considered,
    \item Papers should be written in English,
    \item Only journal and conference papers are considered,
    \item Review papers are not included.
\end{itemize}

\subsection{Data collection}
The search methodology for our systematic review is depicted in \figureabvr \ref{fig:data_collection}. It consists of three major steps, as described below.
\begin{enumerate}
    \item Research findings: In this step, various search engines were used to retrieve relevant articles. The research findings resulted in 588 papers selected for further evaluation.

    \item Elimination: The second step involves applying elimination criteria to refine the selection of papers. We start with duplication remove. The next two steps are the exclusion based on title and abstract screening, and then the full-text screening. 
    \item Final selection: This step presents the outcome of the selection process. 55 articles that matched the inclusion criteria were included in the systematic review. In addition, 16 papers were included based on expert suggestions, bringing the final total to 71 articles. 
    
By following this search methodology, we successfully identified and selected a subset of articles that were most relevant to our research questions and matched the required inclusion criteria. 
\end{enumerate}

\section{Results and findings}\label{sec:sec3}

\subsection{RQ1: classes of deep generative models for ECG, PPG, EEG, and EMG signals}
Based on the selected studies, we identified three classes of deep generative models: 
\begin{enumerate}
    \item Generative adversarial networks (GANs)
    \item  Variational autoencoders (VAEs)
    \item Diffusion models (DMs)
\end{enumerate}

\begin{figure*}[t]
\centering
\includegraphics[width=0.9\textwidth]{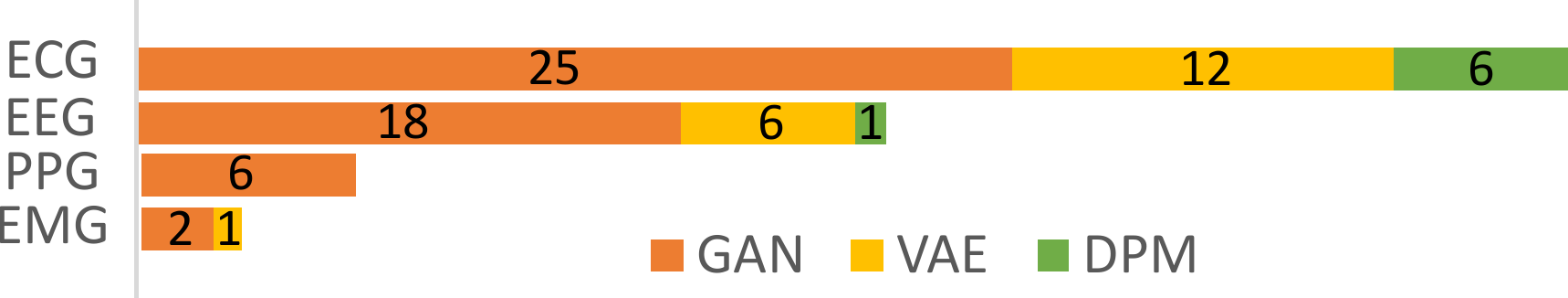}
\caption{Number of papers for each class of deep generative models according to the different considered physiological signals.}
\label{fig:paper_per_signal}
\end{figure*}

\figureabvr \ref{fig:paper_per_signal} presents the detailed number of these deep generative models applied for the different considered physiological signals.
Some studies have focused on applying GANs to multiple types of signals within the same research paper \cite{hazra2020synsiggan,brophy2022deep}, leading to a total number of papers exceeding 71. \tableabvr \ref{tab:papers_per_signals} summarizes the list of selected papers per signal and the employed deep generative models. We can observe that GANs have been widely explored and applied in the domain of physiological signals compared to VAEs and diffusion models, proving their effectiveness. On the other hand, diffusion models, as a relatively recent class of deep generative models, are currently attracting interest and investigation specifically in the context of physiological signals.

\begin{table}[b!]
\centering
\caption{List of papers classified by signal type and deep generative model.}
\begin{tabular}{p{0.06\textwidth} p{0.4\textwidth} p{0.25\textwidth} p{0.25\textwidth}}
\toprule
                  & GANs & VAEs & DMs \\ \midrule
ECG               
& 
\cite{disentanglingnour2022,evolvingSimGANs2022,Golany_Freedman_Radinsky_2021,singh2020new,xu2021ecg,hossain2021ecg,golany2020improving,yang2021proegan,hazra2020synsiggan,mahalanabis2022generative,ye2019ecg,xia2023generative,Golany_Radinsky_2019,abolfazli2022out,adib2022arrhythmia,arnout2019visual,alladi2020augmenting,brophy2022deep,dissanayake2022generalized,lee2022vrnngan,neifar2024leveraging,golany2020simgans,simone2023ecgan,wang2022ecg,zhu2019electrocardiogram}  
& \cite{kuznetsov2020electrocardiogram,sang2022generation,rasmussen2021semi,gyawali2018deep,kuznetsov2021interpretable,gyawali2018learning,lee2021conditional,Robust_Anomaly_2021,pereira2019unsupervised,raza2023anofed,zhu2021ecg,todo2023counterfactual}
& \cite{li2023descod,alcaraz2023diffusion,adib2023synthetic,alcaraz2022diffusion,neifar2024diffecg,s23198328}
\\ \midrule
EEG          
& \cite{an2022auto,brophy2022deep,hazra2020synsiggan,kwon2022novel,du2021multimodal,calhas2020eeg,geng2021auxiliary,lee2021contextual,zhang2021eeg,liu2023eeg,kalashami2022eeg,hu2022e2sgan,song2021improving,truong2019epileptic,xu2022bwgan,zhang2021realizing,biswas2023characterization,prabowo2023advanced}
& \cite{hwaidi2021noise,behrouzi2022graph,bethge2022eeg2vec,ahmed2022examining,li2020latent,tian2023dual}   
& \cite{tosato2023eeg}   \\
\midrule
PPG &  \cite{AnAccurateNonAccelerometer,vo2021p2e,hazra2020synsiggan,brophy2022deep,hwang2022user,sarkar2021cardiogan} & \xmark & \xmark \\\midrule
EMG & \cite{hazra2020synsiggan,mendez2022emg}  &  \cite{olsson2021can}  & \xmark \\
\bottomrule
\end{tabular}
\label{tab:papers_per_signals}
\end{table}

\subsubsection{Generative adversarial networks} 

\begin{figure*}[t]%
\centering
\includegraphics[width=0.8\textwidth]{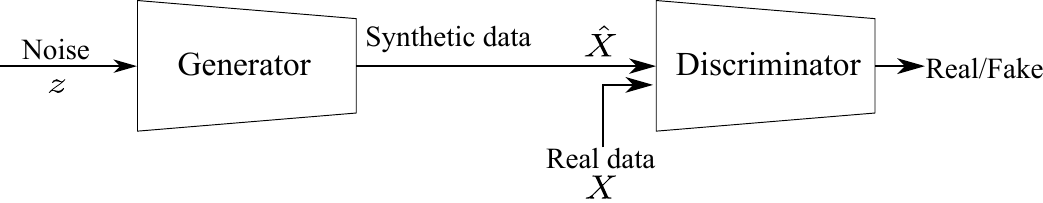}
\caption{Principle of Generative Adversarial Networks.}
\label{fig:gan}
\end{figure*}

Generative Adversarial Networks (GANs), proposed by \cite{goodfellow2020generative}, are the most used class of deep generative models which have gained significant attention in the last years. There were 51 studies from the total selected research that focused on applying GANs with different physiological signals. GANs consist of two neural networks that compete against each other to generate new samples that closely match a particular distribution. \figureabvr \ref{fig:gan} depicts the working principle of GAN. The first network is the generator. Its goal is to synthesize synthetic samples by learning the underlying distribution of the training data. It takes as input random noise and produces synthetic samples similar to real data. The other network is the discriminator. The role of the discriminator is to distinguish between real data and the synthetic data produced by the generator. The goal of the discriminator is to accurately identify the real samples as well as provide feedback to the generator to improve the generated samples. The training of these two networks is formulated as: 
\begin{equation}
\label{ganloss}
\min_{ G } \max_{D}  { \Eb }_{ \Xc} [ \log ( D ( \Xc ) ) ]
+ { \Eb  }_{ \zb }  [ \log ( 1 - D ( G ( \zb ) ) ) ] 
\end{equation}

GANs have been enhanced over time in order to address particular challenges or improve their performance on specific tasks.
\begin{itemize}
    \item Conditional GANs (cGANs): cGANS are extensions of the original GANs that contain additional information in the generation process such as class labels to allow more control over the generated samples. Several of the selected papers \cite{disentanglingnour2022,neifar2024leveraging,zhang2021eeg,wang2022ecg} have proposed a cGAN framework for generating ECG and EEG signals.
    \item Wasserstein GANs (WGANs): WGANs were proposed as a solution to the training instability and mode collapse challenges of GANs by introducing a different objective function based on the Wasserstein distance. For instance, proposed approaches \cite{xu2022bwgan,neifar2024leveraging,disentanglingnour2022,lee2021contextual} are based on WGANs with gradient penalty to improve the training process.
    \item CycleGANs: they are primarily used for unsupervised translation tasks. They are based on learning mappings between two different domains without paired training data. In addition to the adversarial loss, the cycle consistency loss is introduced to create realistic translations and ensure that the translated data can be accurately converted back to the original domain. For example, cycleGAN was used for ECG data translations, imputation, and denoising in \cite{mahalanabis2022generative}. 
    \item Other variants were employed such as Auxiliary Classifier GAN (ACGAN) in \citep{geng2021auxiliary}, Deep Convolutional GAN (DCGAN) in \citep{truong2019epileptic}, Least Square GAN (LSGAN) in \citep{singh2020new}.

\end{itemize}

\begin{figure*}[t]
\centering
\includegraphics[width=0.9\textwidth]{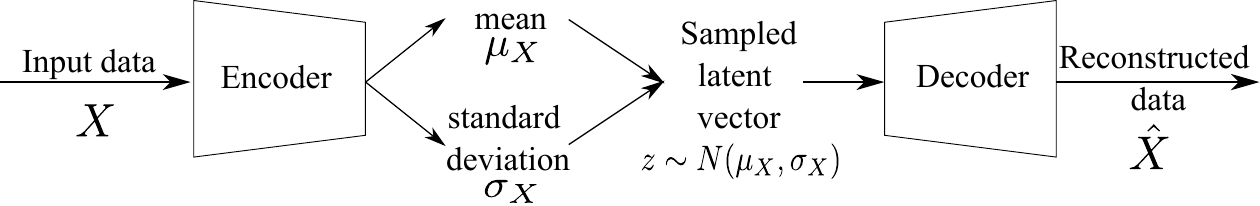}
\caption{Principle of Variational Autoencoders.}
\label{fig:vae}
\end{figure*}
\subsubsection{Variational autoencoders:}
VAEs, proposed by Kingma \etal\cite{kingma2013auto}, are widely used in various domains as a class of deep generative models. The main concept behind VAEs is to transform input data to a low-dimensional latent space representation. \figureabvr \ref{fig:vae} presents the principle of VAE. The VAE is composed of two neural networks. The first network is called the encoder. This network maps the input data to a latent space, often assumed to be a Gaussian distribution with a learned mean and variance. The other network is the decoder. This network takes a sample from the latent space distribution and reconstructs the original input data. The decoder's goal is to produce a reconstructed sample closely similar to the input data. During the training step, the parameters of the encoder and decoder are optimized in order to minimize the reconstruction error. Additionally, a regularization term called the Kullback-Leibler (KL) divergence is introduced to ensure that the learned latent space distribution is similar to a standard Gaussian distribution. The training of the basic VAE is formulated as: 
\begin{equation}
\label{vaeloss}
Loss  = \| \Xc - \hat{\Xc} \|^2 + KL[N(\mu_{\Xc},\sigma_{\Xc}),N(0,1)]
\end{equation}
Several variants of the VAE have been proposed to enhance its performance and address specific challenges:
\begin{itemize}
    \item Conditional VAEs (cVAEs): cVAEs are an extension of VAEs where conditional information are incorporated during both the encoding and decoding processes, allowing the generation of samples conditioned on specific input conditions. For example, in \cite{sang2022generation,bethge2022eeg2vec} conditional VAEs were proposed for 12-lead ECG generation and learning EEG representations.
    \item Variational Graph AutoEncoders (VGAEs): VGAEs are designed for unsupervised learning on graph-structured data. In \cite{behrouzi2022graph}, a VGAE is proposed to extract nodal features of EEG functional connectivity.
    \item Other variants were used in the selected studies such as Convolutional VAEs (CNN-VAEs) in \cite{ahmed2022examining} and Variational Recurrent Autoencoders (VRAEs) in \cite{zhu2021ecg}
\end{itemize}

\begin{figure*}[t]
\centering
\includegraphics[width=0.65\textwidth]{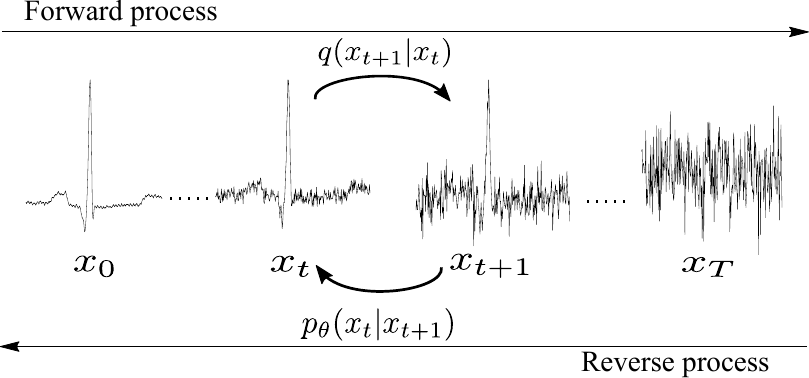}
\caption{Principle of Diffusion models. Here, an ECG signal generation by denoising diffusion probabilistic model is taken as an example for illustration purposes.}
\label{fig:diffusion}
\end{figure*}

\subsubsection{Diffusion models:}
Diffusion models are a rising class of deep generative models with different method for modeling data distributions. In contrast to GANS and VAEs, diffusion models are based on employing a sequence of transformations on the input distribution. \figureabvr \ref{fig:diffusion} presents the principle of diffusion model. The basic concept behind diffusion models is to disturb the input data by sequentially adding noise. Then a reverse process is applied to transform the noise distribution back into the desired data distribution.
Current selected studies on diffusion models are mostly based on one type of diffusion models:
\begin{itemize}
    \item Denoising diffusion probabilistic models (DDPMs) \cite{ho2020denoising}: DDPMS are a specific class of diffusion models based on two Markov chains: forward and reverse diffusion processes. During the forward process, a Gaussian noise $\epsilon$ is incrementally introduced to the input data $x_0$ from the real distribution $D$ over a number of steps $T$ until converging to a standard Gaussian distribution. In the reverse process, a learned model is trained to remove the noise and recover the original data by learning the inverse mapping. The training process (\ref{eq:ddpm}) involves optimizing the model parameters to minimize the reconstruction error between the denoised output data and the original data. 
 \begin{align}
  \label{eq:ddpm}   \min _ {\theta } \; \Eb_{x_0 \sim P, \epsilon \sim N (0, 1), t \sim U(0,T)} \lVert \epsilon - \epsilon_\theta(\sqrt{\alpha_t}x_0 +(1-\alpha_t)\epsilon,t) \rVert _2^2
 \end{align} where $\epsilon_\theta$ is the denoising function that estimates the noise $\epsilon$ introduced to $x_t$.
    This variant of diffusion models was used in different studies, such as \citep{neifar2024diffecg,ho2020denoising,tosato2023eeg}
    
    \item Other variants of diffusion models were proposed such as Score-Based Generative Models (SGMs) \cite{song2019generative}. SGMs focus on learning the score function of the data distribution, which represents the gradient of the log-density function. This variant has not been employed in any selected studies.

\end{itemize}

\subsection{RQ2: application of deep generative models}
Deep generative models have been employed in various applications that have considerably contributed to advancements in the medical field. 
The main considered applications of GANs, VAEs, and DMs in the selected papers are: 
\begin{multicols}{2}
\begin{itemize}
    \item Data augmentation
    \item Denoising
    \item Forecasting
    \item Imputation
    \item Modality transfer 
    \item Anomaly detection
\end{itemize}
\end{multicols}

\begin{figure*}[t]
\centering
\includegraphics[width=0.8\textwidth]{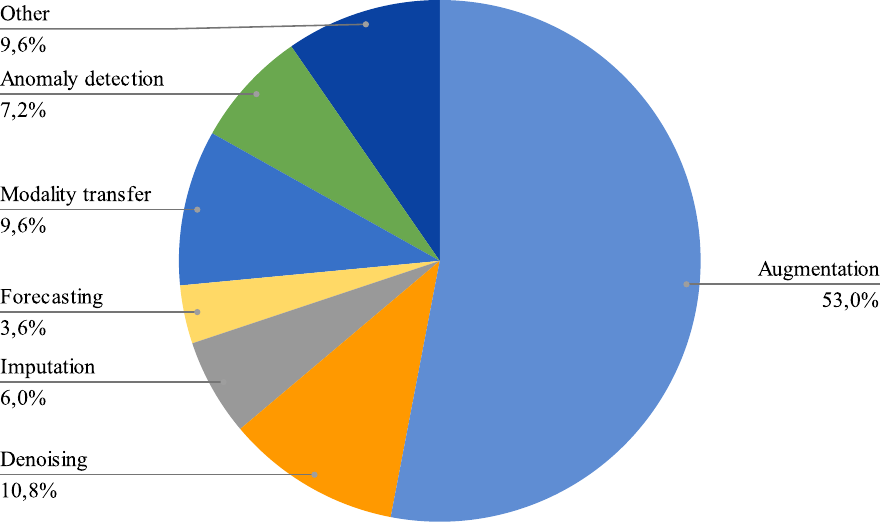}
\caption{Distribution of literature per application.}
\label{fig:task_distribution}
\end{figure*}
\figureabvr\ref{fig:task_distribution} represents the distribution of literature per application. \tableabvr \ref{tab:methods_summary} summarizes the list of papers focusing on the above various tasks classified by signal type and deep generative model approach.

\subsubsection{Data augmentation}
Deep generative models are commonly used to augment medical datasets for various purposes, in particular when using small and imbalanced datasets. Medical datasets frequently suffer from limited training data, which can significantly impact the effectiveness of deep learning models. However, these datasets can be augmented by using deep generative models. Generating synthetic samples will result in a larger and more varied training set, enabling deep learning models to accurately learn the representation of the principal patterns seen in the medical data. Furthermore, collecting positive data related to some medical emergencies (\eg epileptic seizures) can be challenging, mainly due to the unexpected nature of these events. Medical emergencies can happen suddenly without prior warning, which makes it challenging to collect a sufficient amount of positive instances leading to imbalanced datasets. By generating synthetic examples of the underrepresented conditions, these datasets can be balanced to enhance deep models' performance. \tableabvr \ref{tab:methods_summary} (\cf data augmentation rows) summarizes the considered studies that are mainly concerned with using physiological signals generation.

\FloatBarrier
\begin{sidewaystable}[]
    \caption{List of papers focusing on the various tasks classified by signal type and deep generative model approach.}
    \centering
  \begin{tabular}{p{0.01\textwidth} | p{0.05\textwidth} |  p{0.35\textwidth} p{0.2\textwidth} p{0.1\textwidth} p{0.1\textwidth}}
\cmidrule[1pt]{3-6}
\multicolumn{1}{c}{} & \multicolumn{1}{c}{} & ECG & EEG  & PPG & EMG\\
\cmidrule[1pt]{1-6}
\makecell[c]{\multirow{3}{*}[10pt]{\rotatebox[origin=c]{90}{\textls[-50]{\footnotesize Data augmentation}}}} & GANs    & \cite{disentanglingnour2022,evolvingSimGANs2022,Golany_Freedman_Radinsky_2021,hossain2021ecg,golany2020improving,yang2021proegan,hazra2020synsiggan,mahalanabis2022generative,ye2019ecg,xia2023generative,Golany_Radinsky_2019,adib2022arrhythmia,alladi2020augmenting,brophy2022deep,dissanayake2022generalized,lee2022vrnngan,neifar2024leveraging,golany2020simgans,simone2023ecgan,zhu2019electrocardiogram}  & \cite{hazra2020synsiggan,geng2021auxiliary,zhang2021eeg,liu2023eeg,kalashami2022eeg,song2021improving,xu2022bwgan,biswas2023characterization,prabowo2023advanced}  & \cite{hazra2020synsiggan,hwang2022user}  & \cite{hazra2020synsiggan,mendez2022emg}  \\ \cmidrule[0.1pt]{2-6}
& VAEs   & \cite{kuznetsov2020electrocardiogram,sang2022generation,kuznetsov2021interpretable,lee2021conditional}
&  \cite{tian2023dual} &   \xmark  & \cite{olsson2021can}  \\ \cmidrule[0.1pt]{2-6}
& DMs    &  \cite{alcaraz2023diffusion,adib2023synthetic,neifar2024diffecg,s23198328}  & \cite{tosato2023eeg}    &   \xmark  & \xmark  \\
\toprule[1pt]
\multirow{3}{*}[-5pt]{\rotatebox[origin=c]{90}{\textls[-60]{\footnotesize Denoising}}} & GANs &  \cite{singh2020new,xu2021ecg,mahalanabis2022generative,wang2022ecg}  &   \cite{an2022auto,brophy2022deep}   &   \cite{AnAccurateNonAccelerometer}  &  \xmark  \\\cmidrule[0.1pt]{2-6}
& VAEs   &  \xmark  &  \cite{hwaidi2021noise}   &  \xmark   &  \xmark   \\\cmidrule[0.1pt]{2-6}
& DMs    &    \cite{li2023descod}  &    \xmark  &    \xmark  &   \xmark    \\
\toprule[1pt]
\multirow{3}{*}[-5pt]{\rotatebox[origin=c]{90}{\textls[-40]{\footnotesize Imputation \!\!}}} & GANs &   \cite{mahalanabis2022generative}  & \cite{lee2021contextual}    &    \xmark &   \xmark \\\cmidrule[0.1pt]{2-6}
& VAEs   &  \cite{lee2021conditional}  &  \xmark  &  \xmark   &  \xmark   \\\cmidrule[0.1pt]{2-6}
& DMs    &  \cite{alcaraz2022diffusion,neifar2024diffecg}  &    \xmark  &    \xmark  &   \xmark    \\
\toprule[1pt]
\multirow{3}{*}[-4pt]{\rotatebox[origin=c]{90}{\footnotesize  \textls[-50]{Forecasting}}} & GANs & \xmark & \cite{truong2019epileptic}    &   \xmark  &    \xmark  \\\cmidrule[0.1pt]{2-6}
& VAEs   &  \xmark  &  \xmark  &  \xmark   &  \xmark   \\\cmidrule[0.1pt]{2-6}
& DMs    &  \cite{neifar2024diffecg,alcaraz2022diffusion}  &   \xmark  & \xmark    &   \xmark \\
\toprule[1pt]

\multirow{4}{*}[-5pt]{\rotatebox[origin=c]{90}{\footnotesize \textls[-50]{Modality transfer\!\!\!\!}}} & GANs &   \xmark &   \cite{kwon2022novel,du2021multimodal,calhas2020eeg}  &  \cite{sarkar2021cardiogan,vo2021p2e,brophy2022deep} & \xmark \\
 &  &  &   \cite{hu2022e2sgan,zhang2021realizing}  &   &  \\\cmidrule[0.1pt]{2-6}
& VAEs   &  \xmark  &  \xmark  &  \xmark   &  \xmark   \\\cmidrule[0.1pt]{2-6}
& DMs    &  \xmark    &   \xmark  & \xmark    &   \xmark \\
\toprule
\multirow{4}{*}{\rotatebox[origin=c]{90}{  \footnotesize \textls[-60]{Anomaly detection\!\!\!\!} }} & GANs & \cite{abolfazli2022out}  & \xmark    &   \xmark  & \xmark   \\\cmidrule[0.1pt]{2-6}
& VAEs   & \cite{rasmussen2021semi}, \cite{Robust_Anomaly_2021} , \cite{pereira2019unsupervised}, \cite{todo2023counterfactual} &  \xmark  &  \xmark   &  \xmark   \\
&  & \cite{raza2023anofed}   &  &  &   \\\cmidrule[0.1pt]{2-6}
& DMs    &  \xmark    &   \xmark  & \xmark    &   \xmark \\
\toprule[1pt]

\end{tabular}
    \label{tab:methods_summary}
\end{sidewaystable}
\FloatBarrier

Many studies (28.16\%) focused on applying GAN for ECG generation. For instance, several approaches were proposed in order to balance the different arrhythmia classes, by generating samples from these minor classes \cite{golany2020improving,golany2020simgans,neifar2024leveraging}. In addition, VAEs were also employed to ECG generation. For example, Sang \etal\cite{sang2022generation} used a conditional VAE to generate 12-lead ECG.

\subsubsection{Denoising}
Physiological signals can be distorted by numerous types of noise and artifacts. Several noise sources that may affect signal quality can be detected, including baseline wander, muscle artifacts, and environmental noise. Deep generative models are widely employed for signals denoising purpose. They have shown promising results in removing undesired noise and improving physiological signals quality, resulting in more accurate analysis and diagnosis. The considered studies that deal with signals denoising are regrouped in \tableabvr \ref{tab:methods_summary} (\cf denoising rows).
For example, Afandizadeh \etal\cite{AnAccurateNonAccelerometer} proposed a CycleGAN framework for PPG denoising particular from motion artifacts. Furthermore, Li \etal\cite{li2023descod} proposed a conditional score-based diffusion framework for removing baseline wander and noise in ECG signals. 

\subsubsection{Imputation}
Missing data represents a significant challenge in the analysis of physiological signals. It could be caused by various factors such as sensor malfunction or data transmission errors. This missing data can limit the effectiveness of the analysis and interpretation of the signals. However, deep generative models have emerged as an effective solution for handling missing values problems in physiological signals. \tableabvr \ref{tab:methods_summary} (\cf imputation rows) summarizes the corresponding research for physiological signals imputation. Alcaraz \etal\cite{alcaraz2022diffusion} proposed a novel solution for ECG imputation by using conditional diffusion models and structured state space models. Furthermore, Mahalanabis employed a CycleGAN framework in her thesis for ECG imputation \cite{mahalanabis2022generative}. In this approach, the author used Long Short-Term Memory (LSTM) for the generator and discriminator. The Wasserstein loss was used to train the CycleGAN model.

\subsubsection{Forecasting}
Signal forecasting remains a significant tool in health monitoring as it allows the prediction of future changes in a patient's state, allowing for appropriate decisions and timely interventions. Deep generative models are commonly used to make accurate predictions and detect variations of future signal values. They have demonstrated their ability to capture the different patterns inherent in physiological signals and to learn their temporal dependencies. \tableabvr \ref{tab:methods_summary} (\cf forecasting rows) provides an overview of the primary studies for physiological signals forecasting. 
For example, Neifar \etal\cite{neifar2024diffecg} presented a novel framework based on the denoising diffusion probabilistic models for synthesizing ECG. In this approach, three scenarios are covered including full heartbeat forecasting. In addition, two additional conditions related to the prior shape of the ECG are employed to guide the reverse process in cases of imputation or forecasting, ensuring realistic and accurate synthetic ECG signals.

\subsubsection{Modality transfer}
\rev{The increasing importance of multimodal medical data has been considered, particularly with deep generative models for integrating and synthesizing such data. Indeed, several studies have investigated multimodal data generation, such as the use of diffusion models for generating CT and MRI modalities \mbox{\citep{khader2023denoising}}. However, within the scope of physiological signals, we observed that studies on multimodal data generation remain very limited despite a couple of studies focusing on a related topic, the modality transfer}. Modality transfer is an effective technique with several applications in the medical field. It can be used for improving signal analysis, combining information obtained from different modalities to a more accurate diagnosis of physiological states, or overcoming data limitation problems with a particular modality. Employing deep generative models for this task contributes significantly to better understanding the physiological systems and enhancing the disease diagnosis. The primary studies that have focused on modality transfer are presented in \tableabvr \ref{tab:methods_summary} (\cf modality transfer rows). For example, Sarkar \etal\cite{sarkar2021cardiogan} proposed a GAN framework called CardioGAN based on the CycleGAN architecture to generate ECG from PPG signals. 

\subsubsection{Anomaly Detection}
Detecting anomalies in physiological signals is crucial as it can help identify potential health issues and monitor patient conditions. Deep generative models can be widely employed to identify abnormal patterns in physiological signals, helping in the detection of various health conditions. These models can effectively identify deviations from the expected patterns by learning the underlying normal distribution of data, enabling early diseases identification and diagnosis. The primary studies that have focused on anomaly detection are presented in \tableabvr \ref{tab:methods_summary} (\cf anomaly detection rows). For instance, Rasmussen \etal\cite{rasmussen2021semi} proposed an approach that combines an unsupervised VAE with a supervised classifier to differentiate between atrial fibrillation and non-atrial fibrillation.

\subsubsection{Other applications}
GANs have also been successfully applied to translate between different classes of the same physiological signals. This could be useful for various problems, such as the limited volume of signals and the lack of diversity in profiles or conditions. For example, a GAN model called RhythmGAN for translating between different classes of ECG profiles for the same individual was introduced in \cite{alladi2020augmenting}. 
VAEs have indeed been applied to various other applications. Gyawali \etal\cite{gyawali2018deep} proposed a VAE that is utilized to disentangle and identify unobserved confounding factors in ECG signals. A VAE model, presented by Gyawali \etal\cite{gyawali2018learning}, is employed to disentangle the variations present in ECG individual data. Another application of VAE, discussed by Zhu \etal\cite{zhu2021ecg}, is the learning of a significant representation of ECG signals which will be used for various tasks, including clustering similar ECG patterns. Other additional uses of VAEs in the context of EEG data are proposed. They have been employed to extract nodal features that capture the functional connectivity of the brain based on EEG data \citep{behrouzi2022graph}. They are also used for dimensionality reduction \cite{ahmed2022examining} and learning latent factors or representations that capture meaningful features in EEG data \cite{li2020latent}.
Also, a conditional VAE-based framework, called EEG2Vec, was proposed by Bethge \etal\cite{bethge2022eeg2vec}, to learn generative-discriminative representations from EEG that could be employed for affective state estimation, emotion generation as well as synthesis of subject-specific multi-channel EEG signals.

\subsection{RQ3: main challenges associated with using deep generative models for ECG, EEG, PPG, and EMG signals}
Several major challenges are faced when applying deep generative models to physiological signals. The most commonly faced problem with GANs and VAEs is training instability, whereas diffusion models provide a more stable training process. The reason behind training instability with GANs is the adversarial nature of their training where the generator and discriminator networks compete in a min-max game. The generator attempts to synthesize realistic samples to fool the discriminator, whereas the discriminator tries to accurately identify real and generated samples. This sensitive balance can result in instability problems as mode collapse or vanishing gradients. Mode collapse occurs when the generator is unable to capture the full diversity of the data distribution, leading to limited variations in the generated samples. While vanishing gradients, which occur when the discriminator gets better during training, can limit the learning and make it difficult to train the generator network successfully. For example, the proposed approaches in \cite{xu2022bwgan,neifar2024leveraging,adib2022arrhythmia,disentanglingnour2022,song2021improving,lee2021contextual,zhang2021eeg} were based on using WGAN for training stability. For ECG denoising, a LSGAN framework was proposed by Singh \etal\cite{singh2020new}. To stabilize the GAN training process, the original cross-entropy loss function was changed by the least-square function. Another technique was proposed by Ye \etal\cite{ye2019ecg} to address the instability during training through the use of policy gradient in reinforcement learning with SeqGAN.

Furthermore, VAEs can suffer from training instability. VAEs try to optimize a compromise between the two losses in their objective functions: the reconstruction and the regularization terms with the aim of learning a significant latent representation of the data. However, finding the optimal balance between these terms can be difficult. Overfitting can occur as a result of inadequate regularization in which the model succeeds in memorizing the training data but fails to generalize well to new samples. On the other hand, excessive regularization may result in blurry reconstructions or inadequate diversity in the generated samples. 

On the other hand, diffusion models provide more stability during training. Diffusion models progressively transform the simple initial distribution into the target distribution by iteratively denoising the data through a step-by-step process, Contrary to GANs and VAEs, no adversarial training or complex regularization is needed. This simplicity results in more stable training and higher data quality. However, it is essential to highlight that diffusion models have their specific challenges. Achieving a balance between data quality and training stability requires making appropriate design choices, such as selecting appropriate diffusion steps and noise schedules. Moreover, due to the iterative nature of the training of diffusion models, it can be more computationally complex than the training of GANs and VAEs, needing additional time and resources.

Another challenge when applying deep generative models to ECG, EEG, PPG, and EMG is their complex dynamics and nature. These challenges result from the complex variations and dynamics present within physiological signals, as well as their high-dimensionality and inter-/intra-individual variability. \rev{Moreover, handling multimodal physiological data introduces additional challenges. Synchronizing different modalities, such as EEG and ECG, adds complexity, requiring coherent data representation and precise alignment across various physiological sources.} Furthermore, the presence of multiple leads further increases the modeling complexity of these signals, since each lead records a distinct aspect of the physiological activity. To address these challenges, the development of advanced deep generative models specially designed to overcome the complex dynamics of physiological signals is required. Recent selected GAN-based studies \cite{Golany_Radinsky_2019,golany2020simgans,evolvingSimGANs2022,Golany_Freedman_Radinsky_2021,disentanglingnour2022,simone2023ecgan,neifar2024leveraging} have focused on integrating customized prior knowledge of ECG dynamics and patterns into the generation process. Leveraging customized prior knowledge involves incorporating domain-specific information such as specific patterns of ECG signals (P, QRS, T waves) into the generative process. By using this knowledge, the generation will be more guided while maintaining the dynamics and patterns observed in real ECG data. For instance, Golany \etal\cite{Golany_Freedman_Radinsky_2021} proposed to incorporate physical considerations related to ECG signals as supplementary input into the generation process. In addition, Neifar \etal\cite{neifar2024leveraging} introduced a novel prior knowledge modeling about ECG shape and dynamics by integrating statistical shape modeling. Indeed, by leveraging statistical shape model the GAN will be able to encode prior knowledge about the shape variations observed in ECG signals. This prior knowledge provides useful guidance to the generation process, enabling the GAN to generate ECG signals with realistic shape characteristics and correspond to the expected variations.

\subsection{RQ4: commonly used protocols evaluation for assessing the performance of deep generative models}
We have identified two protocols of evaluation from the selected studies: A qualitative and quantitative evaluation. The qualitative evaluation consists of visual inspection and assessment for coherence, fidelity, and consistency of the deep generative models' outputs. More than 60\% of the selected studies have evaluated the quality of the used deep generative models outputs visually. During this evaluation, real and synthetic signals are visually compared with the goal of looking for similarities, differences, and overall coherence. For example, in signal augmentation task, the studies \cite{hossain2021ecg,yang2021proegan,mahalanabis2022generative,xia2023generative,zhang2021eeg,kuznetsov2020electrocardiogram,alcaraz2023diffusion} compared synthetic signals with real signals. Similarly, in denoising tasks, \cite{xu2021ecg} compared denoised signals with real signals. Furthermore, experts in the medical field such as cardiologists may contribute to the qualitative evaluation by providing their domain-specific knowledge and expertise for assessing the coherence and fidelity of the generated signals \cite{neifar2024leveraging,adib2023synthetic}. In addition to visual comparisons, other techniques such as t-SNE (t-Distributed Stochastic Neighbor Embedding), PCA (Principal Component Analysis), and UMAP (Uniform Manifold Approximation and Projection) have been employed to compare the distributions of real and synthetic signals in lower-dimensional spaces. For example, research proposed in \cite{lee2022vrnngan,neifar2024diffecg,simone2023ecgan,geng2021auxiliary,xu2022bwgan} employed t-SNE and UMAP to visualize the distribution of real and synthetic samples in a lower-dimensional space. Additionally, Kalashami \etal\cite{kalashami2022eeg} used PCA to analyze the extracted features of real and fake EEG signals. 

\begin{table}[bh!]
\caption{The used evaluation metrics organized by task.}
\label{tab:metrics}
\begin{tabular}{l p{0.73\textwidth}}

\toprule
{\textbf{Tasks}} & {\textbf{Evaluation metrics}}\\ 
\midrule
Augmentation               
& Euclidean Distance (ED),
Dynamic Time Warping (DTW), 
Pearson's Correlation Coefficient (PCC),
Kullback Leibler Divergence (KLD), 
Root Mean Square Error (RMSE),
Percent Root Mean Square Difference (PRD),
Mean Absolute Error (MAE), 
Frechet distance (FD), 
Kernel Maximum Mean Difference (KMMD),
Relative entropy (RE), 
Time-warp Edit Distance (TWED),
Soft-DTW,
Maximum Mean Discrepancy (MMD), 
Multivariate DTW (MVDTW), 
Mean Squared Error (MSE), 
Earth Mover's Distance (EMD),
Inception Score (IS), 
Frechet Inception Distance (FID),
Wasserstein distance (WD),
Chamfer distance (CD),
Jensen–Shannon divergence (JSD),
Structural Similarity Index (SSIM),
Cross-correlation coefficient,
Normalized Mean Squared Error (NRMSE).
\\ \midrule
Denoising               
& Signal-to-Noise Ratio (SNR),
signal-to-noise ratio improvement (SNRimp),
PRD, MSE, RMSE, 
Sum of the square of the distances (SSD), 
Absolute maximum distance (MAD), 
mean correlation,
peak-to-peak error (PPE),
Cosine similarity.
\\ \midrule
Imputation            
& mean absolute percent error (MAPE),
RMSE, IS, FID, MAE, Continuous Ranked Probability Score (CRPS), MSE, CD, EMD, MMD.
 \\   \midrule
Forecasting               
& MSE, RMSE, MAE, FID, CD, EMD, MMD, CRPS.
\\ \midrule
Modality transfer               
& MAE, RMSE, PRD, FD, PCC, DTW.
\\ \midrule
Anomaly detection               
& Precision, Recall, F1-score, Accuracy.
\\ \bottomrule

\end{tabular}
\end{table}

\begin{figure*}[t]%
\centering
\includegraphics[width=0.9\textwidth]{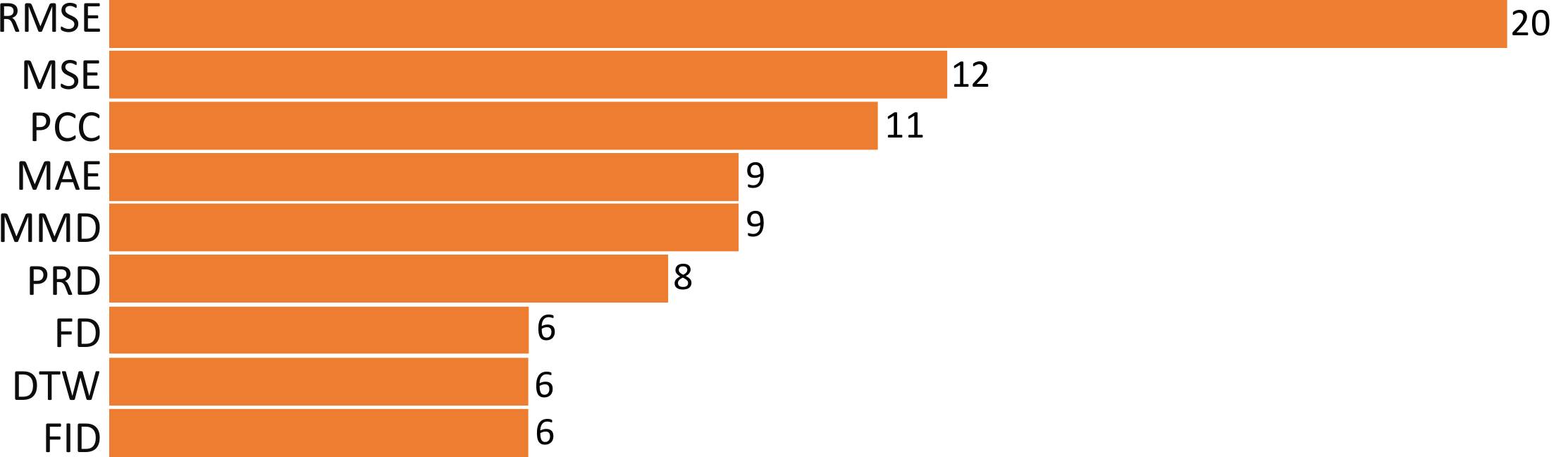}
\caption{The most used metrics to evaluate deep generative models for ECG, EEG, PPG, and EMG signals.}
\label{fig:metrics}
\end{figure*}

On the other hand, the quantitative evaluation involves the use of distance or statistical evaluation metrics. These metrics provide quantitative indications of similarity, and dissimilarity between the deep generative models outputs and real data. \tableabvr \ref{tab:metrics} summarizes the used metrics in the primary studies in the different applications. \figureabvr \ref{fig:metrics} depicts the most used evaluation metrics for the different tasks for ECG, EEG, PPG and EMG. The RMSE is used to quantify the stability between signals. While the MSE is calculated to measure the average squared difference. However, the PCC is employed to assess the relationship between signals. The MAE provides the average of the absolute differences. Whereas, the MMD calculates the dissimilarity between signals. The PRD is used to measure the distortion between signals. On the other hand, the FD is calculated to measure the similarity between signals by considering the location and order of the data points. The similarity between two signals is also measured by the DTW metric. On the other hand, the FID measures the similarity of data distributions. 
\tableabvr \ref{tab:formula_metrics} summarizes the formula of the most used metrics.

\begin{table}[ph!]
\caption{The formula of the most used evaluation metrics.}
\label{tab:formula_metrics}
\begin{tabular}{l p{0.8\textwidth}}

\toprule
{\textbf{Metrics}} & {\textbf{Formula}}\\ 
\midrule
RMSE               
& $ \sqrt{\frac{1}{n}\sum_{i=1}^{n}(X_i - \hat{X}_i)^2}$ 
\\ \midrule
MSE               
& $\frac{1}{n} \sum_{i=1}^{n} (X_i - \hat{X}_i)^2$ 
\\ \midrule
PCC            
& $\frac{{\sum_{i=1}^{n} (X_i - \bar{X})(\hat{X}_i - \bar{\hat{X}})}}{{\sqrt{{\sum_{i=1}^{n} (X_i - \bar{X})^2}  {\sum_{i=1}^{n} (\hat{X}_i - \bar{\hat{X}})^2}}}}$ 
 \\   \midrule
MAE               
& $\frac{1}{n} \sum_{i=1}^{n} |X_i - \hat{X}_i|$ 
\\ \midrule
MMD$^2$   &           

$\frac{1}{n(n - 1)} \underset{i \neq j}{\sum} H_{ij}$,  
\newline where $H_{ij} = k(X_i,X_j) + k(\hat{X}_i, \hat{X}_j) - k(X_i, \hat{X}_j) - k(\hat{X}_i, X_j)$  

\\ \midrule
PRD             
& 
$ \text{\rev{100}}  \sqrt{\frac{\sum_{i=1}^{n} (X_i -\hat{X}_i)^2}{\sum_{i=1}^{n} (X_i)^2}}$ 
\\ \midrule
FD               
& $\min_M \underset{(x,\hat{x}) \in M}{\max} e(x_i,\hat{x}_i)$  

\\ \midrule
DTW  &             

$\underset{\pi}{\min}\sqrt{\underset{(i,j) \in \pi}{\sum} d(X_i,\hat{X}_j})^2$
\\ \midrule
FID               
& $\left\lVert \mu_{P} - \mu_{\hat{P}} \right\rVert^2 + \text{Tr}(C_{P} + C_{\hat{P}} - 2\sqrt{C_{P} C_{\hat{P}}})$ 
\\ \midrule 
\\
\multicolumn{2}{l}{\begin{tabular}[c]{@{}l@{}}
\tabitem $X$ and $\hat{X}$ real and generated signals, respectively.  \\ \tabitem $n$ is the length of $X$ and $\hat{X}$. \\
\tabitem $X_i$ is the value of the $i$-th point. \\ \tabitem $\hat{X}_i$ is the generated value of the $i$-th point. \\
\tabitem $\bar{X}$ and $\bar{\hat{X}}$ are the means of the $X$ and $\hat{X}$, respectively. \\ 
\tabitem $k$ is a kernel function. \\
\tabitem \{$x_1$,...,$x_n$\} and \{$\hat{x}_1,...,\hat{x}_n$\} are the sequences of points order of $X$ and $\hat{X}$. \\
\tabitem $M = \{(x_1,\hat{x}_1),...,(x_n,\hat{x}_n)\}$. \\

\tabitem $e$ is the Euclidean distance. \\
\tabitem $d(X_i,\hat{X}_i)$ is the distance between $X_i$ and $\hat{X}_i$. \\
\tabitem $\pi$ is a temporal alignment.\\
\tabitem $P$ and $\hat{P}$ are real and generated distributions. \\  \tabitem $\mu_{P}$ and $\mu_{\hat{P}}$ are the means of $P$ and $\hat{P}$. \\
\tabitem $C_{P}$ and $C_{\hat{P}}$ are the covariance matrices of $P$ and $\hat{P}$. \\ 
\tabitem $\text{Tr}(\cdot)$ is a matrix trace.

\end{tabular} }

\end{tabular}
\end{table}
In the context of data augmentation, these metrics are computed between the synthetic samples and real samples. While, in modality transfer, these metrics could be calculated between the original and translated signals or between the same feature of signals in the transfer task. For example, Sarkar \etal\cite{sarkar2021cardiogan} computed the RMSE, PRD, and FD between reference ECG signals and the reconstructed ECG signals from PPG input. The MAE was also used to compare the extracted heart rate from both the reconstructed ECG and the input PPG. Other metrics in the frequency domain were also explored in \cite{hu2022e2sgan} for the reconstructed signals such as the Hellinger Distance. Furthermore, the effectiveness of deep generative models used in the different discussed applications can also be assessed in relation to particular tasks. For example, classification tasks were conducted using both real and generated data in \cite{neifar2024leveraging,hossain2021ecg,disentanglingnour2022,tosato2023eeg,golany2020simgans,mahalanabis2022generative,adib2022arrhythmia,lee2022vrnngan,geng2021auxiliary,liu2023eeg,lee2021conditional,adib2023synthetic,kwon2022novel,xu2022bwgan}, and metrics such as precision, accuracy, F1 score, recall or cohen’s kappa coefficient can be employed to assess the model's performance. In the context of signal denoising, classification models could be used to test the performance of deep generative models employed for signal denoising \cite{hwaidi2021noise}. Similarly, for anomaly detection tasks, the performance of deep generative models can be assessed based on their abilities to detect unusual patterns in the generated samples, and metrics such as precision, accuracy, recall, or F1 score can be computed to assess their performance. 
\begin{figure*}[t]
\centering
\includegraphics[width=0.9\textwidth]{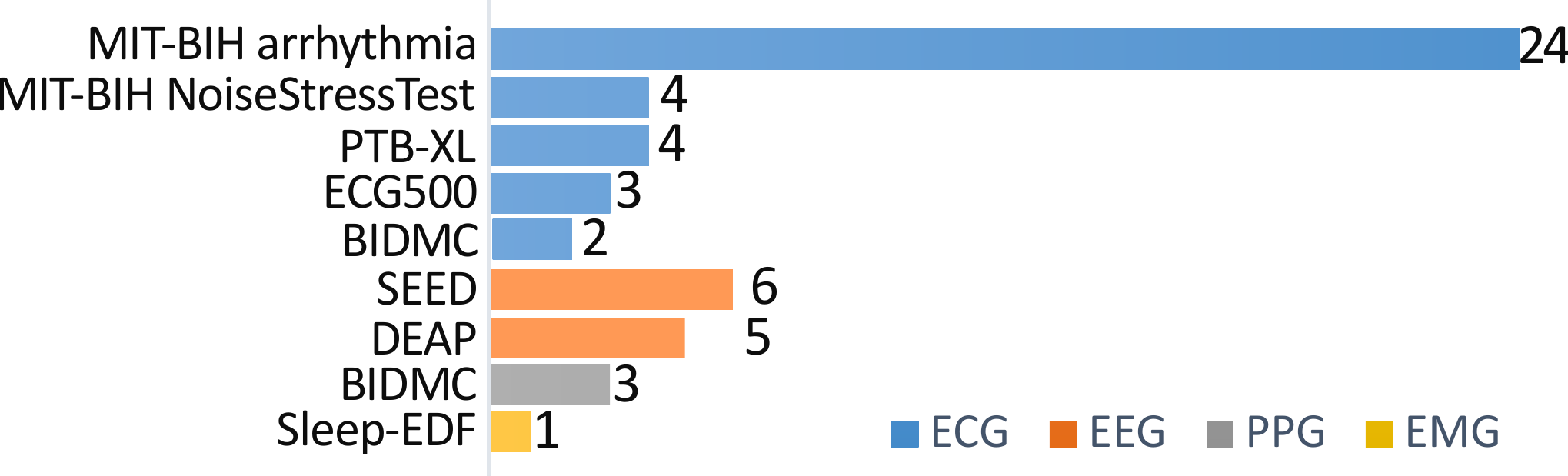}
\caption{Most frequently utilized publicly available databases for ECG, EEG, PPG, and EMG signals generation.}
\label{fig:databases}
\end{figure*}
\begin{table}[ph!]
\caption{Details of the most used publicly available databases.}\label{tab:details_databases}
\scalebox{0.88}{
\begin{tabular}{p{0.09\textwidth} p{1.2\textwidth}}
\toprule
\textbf{Databases} & \; \;   \textbf{Details}  \\
\midrule
MIT- &  \tabitem \footnotesize Most widely used databases for ECG.\\
BIH & \tabitem \footnotesize Signals from both normal heart rhythms and various classes of arrhythmia.  \\
Arrhyth- & \tabitem \footnotesize ECG recordings obtained from 47 half-hour patients. \\
mia &  \tabitem \footnotesize Recordings were digitized at 360 samples. \\
 &\tabitem  \footnotesize Two lead ECG signals. \\
\midrule
BIDMC &  \begin{tabular}[c]{@{}l@{}} \tabitem 53 8-minutes recordings of ECGs, PPGs, and impedance pneumography waveforms. \\
\tabitem \footnotesize Recording was sampled at 125 HZ. \\
\tabitem \footnotesize Three leads of ECG (II, V, AVR).
\end{tabular}    \\
\midrule
MIT-  & \tabitem  \footnotesize Subset of the MIT-BIH Arrhythmia. \\
BIH & \tabitem \footnotesize Evaluating the robustness of arrhythmia detection algorithms in under various\\ 
Noise &  \footnotesize types of noise and artifacts. \\ 
Stress & \tabitem \footnotesize 12 half-hour ECG recordings. \\ 
Test & \tabitem \footnotesize 3 half-hour recordings of noise commonly seen in ECGs (baseline wander, muscle artifact and electrode motion artifact).\\
\midrule
ECG5000 &  \begin{tabular}[c]{@{}l@{}}
\tabitem  \footnotesize Contains 5000 heartbeats from a single patient.  \\
\tabitem \footnotesize  Each heartbeat has a length of 140. \\ 
\tabitem  \footnotesize 5 classes of heartbeats (Normal, R-on-T Premature Ventricular Contraction, Premature \\Ventricular Contraction, Supraventricular Premature beat, and Unclassified Beat).
\end{tabular}     \\

\midrule
PTB-XL   &  \begin{tabular}[c]{@{}l@{}}

\tabitem  \footnotesize 21837 ECG recordings of 10-second length. \\ 
\tabitem  \footnotesize 12-lead ECG annotated by 2 cardiologists. \\
\tabitem  \footnotesize ECG signals with sampling frequency of 100 and 500 Hz. \\
\tabitem  \footnotesize A wide range of rhythm statements (Myocardial Infarction, Bundle Branch Blocks \etc).
\end{tabular}     \\
\midrule
SEED  &  \begin{tabular}[c]{@{}l@{}}

\tabitem  \footnotesize EEG recordings from 15 participants watching 15 edited video clips. \\
\tabitem \footnotesize EEG recordings were downsampled to 200 Hz. \\
\tabitem \footnotesize Three classes of emotions including positive, neutral, and negative.
\end{tabular}  
  \\
\midrule
DEAP   &  \begin{tabular}[c]{@{}l@{}}

\tabitem  \footnotesize EEG recordings with peripheral physiological waveforms obtained from 32 participants viewing 40 \\ one-minute music video. \\
\tabitem \footnotesize Data was downsampled to 128Hz. \\
\tabitem \footnotesize Each music video was scored by participants between 0 and 9 based on their feelings of \\ valence, arousal, dominance, and liking to each music video.
\end{tabular}    \\
\midrule

Sleep-EDF &  \begin{tabular}[c]{@{}l@{}}

\tabitem  \footnotesize Contains 197 whole-night PolySomnoGraphic sleep recordings including chin EMG, EEG \etc. \\
\tabitem   \footnotesize 82 subjects from sleep cassette and 24 from sleep telemetry. \\
\tabitem \footnotesize EMG signals from sleep cassette and sleep telemetry were respectively sampled at 1Hz and\\ 100 Hz.
\end{tabular}
\\\bottomrule
\end{tabular}
}
\end{table}
\subsection{RQ5: most utilized physiological datasets for deep generative models' evaluation}

Several databases were used to evaluate the effectiveness of deep generative models in the various discussed applications. \figureabvr \ref{fig:databases} shows the most open access used datasets that we identified in the primary studies for ECG. The MIT-BIH arrhythmia database \citep{mitbih} represents one of the most often used databases for ECG signals \cite{disentanglingnour2022,evolvingSimGANs2022,Golany_Freedman_Radinsky_2021,singh2020new,xu2021ecg,hossain2021ecg,golany2020improving,yang2021proegan,hazra2020synsiggan,adib2023synthetic,mahalanabis2022generative,ye2019ecg,xia2023generative,Golany_Radinsky_2019,adib2022arrhythmia,brophy2022deep,dissanayake2022generalized,neifar2024leveraging,Robust_Anomaly_2021,golany2020simgans,simone2023ecgan,wang2022ecg,zhu2019electrocardiogram,neifar2024diffecg}. ECG recordings in this database include signals from both normal heart rhythms and several classes of arrhythmia, thus serving as a useful resource for training arrhythmia detection and classification methods. On the other hand, the MIT-BIH Noise Stress Test database \cite{moody1984noise} was used in 4 papers \cite{li2023descod,xu2021ecg,mahalanabis2022generative,wang2022ecg}. It represents a subset of the MIT-BIH Arrhythmia Database. It was mainly created to assess the robustness of arrhythmia detection methods under various types of noise and artifacts faced in clinical settings. The PTB-XL database has also been used in 4 papers of the primary studies \cite{alcaraz2023diffusion,abolfazli2022out}. It is a large database of 12-lead ECG recordings with a variety of cardiac abnormalities \cite{wagner2020ptb}. The ECG5000 database has appeared in only 3 research \cite{Robust_Anomaly_2021,pereira2019unsupervised,zhu2021ecg}. This dataset, provided by Eamonn Keogh and Yanping Chen, consists of a collection of univariate time series representing ECG heartbeats from normal and pathological conditions, providing a diverse range of physiological patterns for analysis. \rev{Beyond single-modality datasets, few publicly available databases incorporate multimodal physiological signals such as} the BIDMC \citep{BIDMC} that has been used for ECG and PPG research \cite{sarkar2021cardiogan,simone2023ecgan,raza2023anofed,AnAccurateNonAccelerometer,hazra2020synsiggan}. It is composed of a variety of physiological signals in addition to ECGs, PPGs such as arterial blood pressure (ABP) waveforms obtained from a diverse range of participants with different ages, genders, and clinical conditions. Other ECG databases were used such as MIT-BIH Atrial Fibrillation database \cite{rasmussen2021semi}, American Heart Association database \cite{alladi2020augmenting}, European Society of Cardiology ST-T database \cite{alladi2020augmenting}, Creighton University Sustained Ventricular Arrhythmia database \cite{alladi2020augmenting}, EPHNOGRAM database \cite{dissanayake2022generalized} \etc 

For the EEG signals, the SEED and DEAP databases were the most used in the selected studies. They are commonly used in the task of emotion recognition. Finally, the EMG recordings from the Sleep-EDF database were used once for EMG synthesis in \cite{hazra2020synsiggan}. \tableabvr \ref{tab:details_databases} provides more details about these databases.

\section{Discussion and future directions}\label{sec:discussion}
Generative adversarial networks, variational autoencoders, and diffusion models are currently promising methods in the analysis and processing of physiological signals. They have been successfully applied in various tasks including data augmentation, denoising, imputation \etc While these deep generative models have significant advantages, there are still existing challenges mainly the training instability and complex dynamic of physiological signals that require further considerations. Future research should concentrate on numerous important areas, including model enhancement, where further research and development are needed for the enhancement of the performance of deep generative models for physiological signals. In addition, the incorporation of prior knowledge about physiological signals is crucial. Although there have been a few attempts to include prior knowledge into deep generative models, additional exploration in this area should be investigated. \rev{Additionally, a promising research direction is the further exploration of multimodal data, as it can provide a more comprehensive understanding of physiological processes and enhance analytical effectiveness. This should include addressing the challenges associated with their integration.} In addition, the absence of standardized evaluation protocols in particular metrics for deep generative models makes it extremely difficult to assess their performance objectively. Developing a common evaluation protocol is indeed a crucial step. 

\rev{Finally, while this study provides a comprehensive overview of generative models applied to physiological signals, one inherent limitation of the SLR methodology is the lack of comparative experimental results among the included studies. Since not all official implementations of these studies are publicly available, additional effort is required to faithfully reproduce their reported performance. Furthermore, the scope of our systematic literature review covers only English-language sources, which may limit the diversity of research contributions included. However, this focus has allowed for a comprehensive analysis of key research trends and findings.
}

\section{Conclusion}\label{sec:sec4}
In our systematic literature review, we have examined a total of 71 primary studies to explore the use of various deep generative models with ECG, EEG, PPG, and EMG signals. The aim of our review was to address specific research questions and provide an overview of the current deep generative models in addition to their main applications in this domain. We have also examined the fundamentals of GANs, VAEs, and diffusion models, and discussed the challenges associated with employing these models with different physiological signals. Furthermore, we have discussed the evaluation protocols employed in these studies on the most used databases. Finally, we concluded by outlining potential directions for future research. As future work, we aim to extend the scope of our study to cover additional physiological signals. In addition, we intend to provide a more comprehensive synthesis that includes a thorough analysis of the types and architectures of several variants of deep generative models.

\section*{Conflict of interest}
The authors declare that there is no conflict of interest.
\bibliography{main}

\begin{thebibliography}{100}
\expandafter\ifx\csname url\endcsname\relax
  \def\url#1{\texttt{#1}}\fi
\expandafter\ifx\csname urlprefix\endcsname\relax\def\urlprefix{URL }\fi
\expandafter\ifx\csname href\endcsname\relax
  \def\href#1#2{#2} \def\path#1{#1}\fi

\bibitem{lanza2007electrocardiogram}
G.~A. Lanza, The electrocardiogram as a prognostic tool for predicting major
  cardiac events, Progress in cardiovascular diseases 50~(2) (2007) 87--111.
\newblock \href {https://doi.org/https://doi.org/10.1016/j.pcad.2007.03.003}
  {\path{doi:https://doi.org/10.1016/j.pcad.2007.03.003}}.

\bibitem{muhammad2020eeg}
G.~Muhammad, M.~S. Hossain, N.~Kumar, {EEG-based pathology detection for home
  health monitoring}, IEEE Journal on Selected Areas in Communications 39~(2)
  (2020) 603--610.
\newblock \href {https://doi.org/10.1109/JSAC.2020.3020654}
  {\path{doi:10.1109/JSAC.2020.3020654}}.

\bibitem{kloppel2012diagnostic}
S.~Kl{\"o}ppel, A.~Abdulkadir, C.~R. Jack~Jr, N.~Koutsouleris,
  J.~Mour{\~a}o-Miranda, P.~Vemuri, Diagnostic neuroimaging across diseases,
  Neuroimage 61~(2) (2012) 457--463.

\bibitem{risacher2021neuroimaging}
S.~L. Risacher, A.~J. Saykin, Neuroimaging advances in neurologic and
  neurodegenerative diseases, Neurotherapeutics 18~(2) (2021) 659--660.

\bibitem{zhao2019deep}
R.~Zhao, R.~Yan, Z.~Chen, K.~Mao, P.~Wang, R.~X. Gao, Deep learning and its
  applications to machine health monitoring, Mechanical Systems and Signal
  Processing 115 (2019) 213--237.
\newblock \href {https://doi.org/https://doi.org/10.1016/j.ymssp.2018.05.050}
  {\path{doi:https://doi.org/10.1016/j.ymssp.2018.05.050}}.

\bibitem{zhang2020ecg}
J.~Zhang, A.~Liu, M.~Gao, X.~Chen, X.~Zhang, X.~Chen, Ecg-based multi-class
  arrhythmia detection using spatio-temporal attention-based convolutional
  recurrent neural network, Artificial Intelligence in Medicine 106 (2020)
  101856.
\newblock \href {https://doi.org/https://doi.org/10.1016/j.artmed.2020.101856}
  {\path{doi:https://doi.org/10.1016/j.artmed.2020.101856}}.

\bibitem{miao2020continuous}
F.~Miao, B.~Wen, Z.~Hu, G.~Fortino, X.-P. Wang, Z.-D. Liu, M.~Tang, Y.~Li,
  Continuous blood pressure measurement from one-channel electrocardiogram
  signal using deep-learning techniques, Artificial Intelligence in Medicine
  108 (2020) 101919.
\newblock \href {https://doi.org/https://doi.org/10.1016/j.artmed.2020.101919}
  {\path{doi:https://doi.org/10.1016/j.artmed.2020.101919}}.

\bibitem{miften2021new}
F.~S. Miften, M.~Diykh, S.~Abdulla, S.~Siuly, J.~H. Green, R.~C. Deo, {A new
  framework for classification of multi-category hand grasps using EMG
  signals}, Artificial Intelligence in Medicine 112 (2021) 102005.
\newblock \href {https://doi.org/https://doi.org/10.1016/j.artmed.2020.102005}
  {\path{doi:https://doi.org/10.1016/j.artmed.2020.102005}}.

\bibitem{jo2019deep}
T.~Jo, K.~Nho, A.~J. Saykin, Deep learning in alzheimer's disease: diagnostic
  classification and prognostic prediction using neuroimaging data, Frontiers
  in aging neuroscience 11 (2019) 220.

\bibitem{alsubaie2024alzheimer}
M.~G. Alsubaie, S.~Luo, K.~Shaukat, Alzheimer’s disease detection using deep
  learning on neuroimaging: a systematic review, Machine Learning and Knowledge
  Extraction 6~(1) (2024) 464--505.

\bibitem{ISLAM2023109201}
N.~Islam, R.~Khan, S.~K. Das, S.~K. Sarker, M.~M. Islam, M.~Akter, S.~Muyeen,
  {Power transformer health condition evaluation: A deep generative model aided
  intelligent framework}, Electric Power Systems Research 218 (2023) 109201.
\newblock \href {https://doi.org/https://doi.org/10.1016/j.epsr.2023.109201}
  {\path{doi:https://doi.org/10.1016/j.epsr.2023.109201}}.

\bibitem{chen2018deep}
X.~Chen, N.~Pawlowski, M.~Rajchl, B.~Glocker, E.~Konukoglu, {Deep generative
  models in the real-world: An open challenge from medical imaging}, arXiv
  preprint arXiv:1806.05452 (2018).
\newblock \href {https://doi.org/https://doi.org/10.48550/arXiv.1806.05452}
  {\path{doi:https://doi.org/10.48550/arXiv.1806.05452}}.

\bibitem{hwang2023real}
U.~Hwang, S.-W. Kim, D.~Jung, S.~Kim, H.~Lee, S.~W. Seo, J.-K. Seong, S.~Yoon,
  A.~D.~N. Initiative, et~al., Real-world prediction of preclinical
  alzheimer’s disease with a deep generative model, Artificial Intelligence
  in Medicine 144 (2023) 102654.
\newblock \href {https://doi.org/https://doi.org/10.1016/j.artmed.2023.102654}
  {\path{doi:https://doi.org/10.1016/j.artmed.2023.102654}}.

\bibitem{wang2023applications}
R.~Wang, V.~Bashyam, Z.~Yang, F.~Yu, V.~Tassopoulou, S.~S. Chintapalli,
  I.~Skampardoni, L.~P. Sreepada, D.~Sahoo, K.~Nikita, et~al., {Applications of
  generative adversarial networks in neuroimaging and clinical neuroscience},
  Neuroimage 269 (2023) 119898.

\bibitem{laino2022generative}
M.~E. Laino, P.~Cancian, L.~S. Politi, M.~G. Della~Porta, L.~Saba, V.~Savevski,
  Generative adversarial networks in brain imaging: A narrative review, Journal
  of imaging 8~(4) (2022) 83.

\bibitem{huynh2024review}
N.~Huynh, G.~Deshpande, {A review of the applications of generative adversarial
  networks to structural and functional MRI based diagnostic classification of
  brain disorders}, Frontiers in Neuroscience 18 (2024) 1333712.

\bibitem{kazerouni2023diffusion}
A.~Kazerouni, E.~K. Aghdam, M.~Heidari, R.~Azad, M.~Fayyaz, I.~Hacihaliloglu,
  D.~Merhof, Diffusion models in medical imaging: A comprehensive survey,
  Medical image analysis 88 (2023) 102846.

\bibitem{zhao2025diffusion}
H.~Zhao, H.~Lou, L.~Yao, W.~Peng, E.~Adeli, K.~M. Pohl, Y.~Zhang, {Diffusion
  Models for Computational Neuroimaging: A Survey}, arXiv preprint
  arXiv:2502.06552 (2025).

\bibitem{ali2022role}
H.~Ali, M.~R. Biswas, F.~Mohsen, U.~Shah, A.~Alamgir, O.~Mousa, Z.~Shah, {The
  role of generative adversarial networks in brain MRI: a scoping review},
  Insights into imaging 13~(1) (2022) 98.

\bibitem{monachino2023deep}
G.~Monachino, B.~Zanchi, L.~Fiorillo, G.~Conte, A.~Auricchio, A.~Tzovara, F.~D.
  Faraci, {Deep Generative Models: The winning key for large and easily
  accessible ECG datasets?}, Computers in biology and medicine 167 (2023)
  107655.

\bibitem{berger2023generative}
L.~Berger, M.~Haberbusch, F.~Moscato, Generative adversarial networks in
  electrocardiogram synthesis: Recent developments and challenges, Artificial
  Intelligence in Medicine 143 (2023) 102632.

\bibitem{brophy2023generative}
E.~Brophy, Z.~Wang, Q.~She, T.~Ward, {Generative Adversarial Networks in Time
  Series: A Systematic Literature Review}, ACM Computing Surveys 55~(10) (2023)
  1--31.
\newblock \href {https://doi.org/https://doi.org/10.1145/3559540}
  {\path{doi:https://doi.org/10.1145/3559540}}.

\bibitem{zhang2022comprehensive}
D.~Zhang, M.~Ma, L.~Xia, A comprehensive review on gans for time-series
  signals, Neural Computing and Applications 34~(5) (2022) 3551--3571.
\newblock \href {https://doi.org/https://doi.org/10.1007/s00521-022-06888-0}
  {\path{doi:https://doi.org/10.1007/s00521-022-06888-0}}.

\bibitem{hazra2020synsiggan}
D.~Hazra, Y.-C. Byun, {SynSigGAN: Generative Adversarial Networks for Synthetic
  Biomedical Signal Generation}, Biology 9~(12) (2020) 441.
\newblock \href {https://doi.org/https://doi.org/10.3390/biology9120441}
  {\path{doi:https://doi.org/10.3390/biology9120441}}.

\bibitem{brophy2022deep}
E.~Brophy, Deep learning-based signal processing approaches for improved
  tracking of human health and behaviour with wearable sensors, Ph.D. thesis,
  Dublin City University (2022).

\bibitem{disentanglingnour2022}
N.~Neifar, A.~Mdhaffar, A.~Ben-Hamadou, M.~Jmaiel, B.~Freisleben,
  {Disentangling Temporal and Amplitude Variations in ECG Synthesis Using
  Anchored GANs}, in: Proceedings of the 37th ACM/SIGAPP Symposium on Applied
  Computing, SAC '22, Association for Computing Machinery, New York, NY, USA,
  2022, p. 645–652.
\newblock \href {https://doi.org/https://doi.org/10.1145/3477314.3507300}
  {\path{doi:https://doi.org/10.1145/3477314.3507300}}.

\bibitem{evolvingSimGANs2022}
G.~Wang, A.~Thite, R.~Talebi, A.~D'Achille, A.~Mussa, J.~Zutty, {Evolving
  SimGANs to Improve Abnormal Electrocardiogram Classification}, in:
  Proceedings of the Genetic and Evolutionary Computation Conference Companion,
  GECCO '22, Association for Computing Machinery, New York, NY, USA, 2022, p.
  1887–1894.
\newblock \href {https://doi.org/https://doi.org/10.1145/3520304.3534048}
  {\path{doi:https://doi.org/10.1145/3520304.3534048}}.

\bibitem{Golany_Freedman_Radinsky_2021}
T.~Golany, D.~Freedman, K.~Radinsky, {ECG ODE-GAN: Learning Ordinary
  Differential Equations of ECG Dynamics via Generative Adversarial Learning},
  Proceedings of the AAAI Conference on Artificial Intelligence 35~(1) (2021)
  134--141.
\newblock \href {https://doi.org/https://doi.org/10.1609/aaai.v35i1.16086}
  {\path{doi:https://doi.org/10.1609/aaai.v35i1.16086}}.

\bibitem{singh2020new}
P.~Singh, G.~Pradhan, {A New ECG Denoising Framework Using Generative
  Adversarial Network}, IEEE/ACM Transactions on Computational Biology and
  Bioinformatics 18~(2) (2020) 759--764.
\newblock \href {https://doi.org/10.1109/TCBB.2020.2976981}
  {\path{doi:10.1109/TCBB.2020.2976981}}.

\bibitem{xu2021ecg}
B.~Xu, R.~Liu, M.~Shu, X.~Shang, Y.~Wang, {An ECG denoising method based on the
  generative adversarial residual network}, Computational and Mathematical
  Methods in Medicine 2021 (2021) 1--23.
\newblock \href {https://doi.org/https://doi.org/10.1155/2021/5527904}
  {\path{doi:https://doi.org/10.1155/2021/5527904}}.

\bibitem{hossain2021ecg}
K.~F. Hossain, S.~A. Kamran, A.~Tavakkoli, L.~Pan, X.~Ma, S.~Rajasegarar,
  C.~Karmaker, {ECG-Adv-GAN: Detecting ECG Adversarial Examples with
  Conditional Generative Adversarial Networks}, in: 2021 20th IEEE
  International Conference on Machine Learning and Applications (ICMLA), IEEE,
  2021, pp. 50--56.
\newblock \href {https://doi.org/10.1109/ICMLA52953.2021.00016}
  {\path{doi:10.1109/ICMLA52953.2021.00016}}.

\bibitem{golany2020improving}
T.~Golany, G.~Lavee, S.~T. Yarden, K.~Radinsky, {Improving ECG Classification
  Using Generative Adversarial Networks}, in: Proceedings of the AAAI
  Conference on Artificial Intelligence, Vol.~34, 2020a, pp. 13280--13285.
\newblock \href {https://doi.org/https://doi.org/10.1609/aaai.v34i08.7037}
  {\path{doi:https://doi.org/10.1609/aaai.v34i08.7037}}.

\bibitem{yang2021proegan}
H.~Yang, J.~Liu, L.~Zhang, Y.~Li, H.~Zhang, {ProEGAN-MS: A Progressive Growing
  Generative Adversarial Networks for Electrocardiogram Generation}, IEEE
  Access 9 (2021) 52089--52100.
\newblock \href {https://doi.org/10.1109/ACCESS.2021.3069827}
  {\path{doi:10.1109/ACCESS.2021.3069827}}.

\bibitem{mahalanabis2022generative}
A.~Mahalanabis, {Generative Adversarial Networks for ECG generation,
  translation, imputation and denoising}, Master's thesis, University of
  Waterloo (2022).

\bibitem{ye2019ecg}
F.~Ye, F.~Zhu, Y.~Fu, B.~Shen, {ECG Generation With Sequence Generative
  Adversarial Nets Optimized by Policy Gradient}, IEEE Access 7 (2019)
  159369--159378.
\newblock \href {https://doi.org/10.1109/ACCESS.2019.2950383}
  {\path{doi:10.1109/ACCESS.2019.2950383}}.

\bibitem{xia2023generative}
Y.~Xia, Y.~Xu, P.~Chen, J.~Zhang, Y.~Zhang, {Generative adversarial network
  with transformer generator for boosting ECG classification}, Biomedical
  Signal Processing and Control 80 (2023) 104276.
\newblock \href {https://doi.org/https://doi.org/10.1016/j.bspc.2022.104276}
  {\path{doi:https://doi.org/10.1016/j.bspc.2022.104276}}.

\bibitem{Golany_Radinsky_2019}
T.~Golany, K.~Radinsky, {PGANs: Personalized Generative Adversarial Networks
  for ECG Synthesis to Improve Patient-Specific Deep ECG Classification},
  Proceedings of the AAAI Conference on Artificial Intelligence 33 (2019)
  557--564.
\newblock \href {https://doi.org/https://doi.org/10.1609/aaai.v33i01.3301557}
  {\path{doi:https://doi.org/10.1609/aaai.v33i01.3301557}}.

\bibitem{abolfazli2022out}
M.~Abolfazli, M.~Z. Arimani, A.~Host-Madsen, J.~Zhang, A.~Bratincsak,
  {Out-of-Distribution Detection using BiGAN and MDL}, arXiv preprint
  arXiv:2206.01851 (2022).
\newblock \href {https://doi.org/https://doi.org/10.48550/arXiv.2206.01851}
  {\path{doi:https://doi.org/10.48550/arXiv.2206.01851}}.

\bibitem{adib2022arrhythmia}
E.~Adib, F.~Afghah, J.~J. Prevost, {Arrhythmia Classification Using
  CGAN-Augmented ECG Signals}, in: 2022 IEEE International Conference on
  Bioinformatics and Biomedicine (BIBM), IEEE, 2022, pp. 1865--1872.
\newblock \href {https://doi.org/10.1109/BIBM55620.2022.9995088}
  {\path{doi:10.1109/BIBM55620.2022.9995088}}.

\bibitem{arnout2019visual}
H.~Arnout, J.~Kehrer, J.~Bronner, T.~Runkler, {Visual Evaluation of Generative
  Adversarial Networks for Time Series Data }, arXiv preprint arXiv:2001.00062
  (2019).
\newblock \href {https://doi.org/https://doi.org/10.48550/arXiv.2001.00062}
  {\path{doi:https://doi.org/10.48550/arXiv.2001.00062}}.

\bibitem{alladi2020augmenting}
S.~Alladi, {Augmenting Electrocardiogram Datasets using Generative Adversarial
  Networks}, Ph.D. thesis, University of Minnesota (2020).

\bibitem{dissanayake2022generalized}
T.~Dissanayake, T.~Fernando, S.~Denman, S.~Sridharan, C.~Fookes, {Generalized
  Generative Deep Learning Models for Biosignal Synthesis and Modality
  Transfer}, IEEE Journal of Biomedical and Health Informatics (2022).
\newblock \href {https://doi.org/10.1109/JBHI.2022.3223777}
  {\path{doi:10.1109/JBHI.2022.3223777}}.

\bibitem{lee2022vrnngan}
J.~Lee, Vrnngan: A recurrent vae-gan framework for synthetic time-series
  generation, Ph.D. thesis, University of Toronto (Canada) (2022).

\bibitem{neifar2024leveraging}
N.~Neifar, A.~Ben-Hamadou, A.~Mdhaffar, M.~Jmaiel, B.~Freisleben, Leveraging
  statistical shape priors in gan-based ecg synthesis, IEEE Access (2024).
\newblock \href {https://doi.org/https://doi.org/10.1109/ACCESS.2024.3373724}
  {\path{doi:https://doi.org/10.1109/ACCESS.2024.3373724}}.

\bibitem{golany2020simgans}
T.~Golany, K.~Radinsky, D.~Freedman, {SimGANs: Simulator-Based Generative
  Adversarial Networks for ECG Synthesis to Improve Deep ECG Classification},
  in: International Conference on Machine Learning, PMLR, 2020b, pp.
  3597--3606.

\bibitem{simone2023ecgan}
L.~Simone, D.~Bacciu, {ECGAN: Self-supervised generative adversarial network
  for electrocardiography}, in: Artificial Intelligence in Medicine, Springer
  Nature Switzerland, 2023, p. 276–280.
\newblock \href {https://doi.org/https://doi.org/10.1007/978-3-031-34344-5_33}
  {\path{doi:https://doi.org/10.1007/978-3-031-34344-5_33}}.

\bibitem{wang2022ecg}
X.~Wang, B.~Chen, M.~Zeng, Y.~Wang, H.~Liu, R.~Liu, L.~Tian, X.~Lu, {An ECG
  Signal Denoising Method Using Conditional Generative Adversarial Net}, IEEE
  Journal of Biomedical and Health Informatics 26~(7) (2022) 2929--2940.
\newblock \href {https://doi.org/10.1109/JBHI.2022.3169325}
  {\path{doi:10.1109/JBHI.2022.3169325}}.

\bibitem{zhu2019electrocardiogram}
F.~Zhu, F.~Ye, Y.~Fu, Q.~Liu, B.~Shen, {Electrocardiogram generation with a
  bidirectional LSTM-CNN generative adversarial network}, Scientific reports
  9~(1) (2019) 1--11.
\newblock \href {https://doi.org/https://doi.org/10.1038/s41598-019-42516-z}
  {\path{doi:https://doi.org/10.1038/s41598-019-42516-z}}.

\bibitem{kuznetsov2020electrocardiogram}
V.~Kuznetsov, V.~Moskalenko, N.~Y. Zolotykh, {Electrocardiogram Generation and
  Feature Extraction Using a Variational Autoencoder }, arXiv preprint
  arXiv:2002.00254 (2020).
\newblock \href {https://doi.org/https://doi.org/10.48550/arXiv.2002.00254}
  {\path{doi:https://doi.org/10.48550/arXiv.2002.00254}}.

\bibitem{sang2022generation}
Y.~Sang, M.~Beetz, V.~Grau, {Generation of 12-Lead Electrocardiogram with
  Subject-Specific, Image-Derived Characteristics Using a Conditional
  Variational Autoencoder}, in: 2022 IEEE 19th International Symposium on
  Biomedical Imaging (ISBI), IEEE, 2022, pp. 1--5.
\newblock \href {https://doi.org/10.1109/ISBI52829.2022.9761431}
  {\path{doi:10.1109/ISBI52829.2022.9761431}}.

\bibitem{rasmussen2021semi}
S.~M. Rasmussen, M.~E. Jensen, C.~S. Meyhoff, E.~K. Aasvang, H.~B. S{\l}rensen,
  {Semi-Supervised Analysis of the Electrocardiogram Using Deep Generative
  Models}, in: 2021 43rd Annual International Conference of the IEEE
  Engineering in Medicine \& Biology Society (EMBC), IEEE, 2021, pp.
  1124--1127.
\newblock \href {https://doi.org/10.1109/EMBC46164.2021.9629915}
  {\path{doi:10.1109/EMBC46164.2021.9629915}}.

\bibitem{gyawali2018deep}
P.~K. Gyawali, C.~Knight, S.~Ghimire, B.~M. Horacek, J.~L. Sapp, L.~Wang, {Deep
  Generative Model with Beta Bernoulli Process for Modeling and Learning
  Confounding Factors}, arXiv preprint arXiv:1811.00073 (2018).
\newblock \href {https://doi.org/https://doi.org/10.48550/arXiv.1811.00073}
  {\path{doi:https://doi.org/10.48550/arXiv.1811.00073}}.

\bibitem{kuznetsov2021interpretable}
V.~Kuznetsov, V.~Moskalenko, D.~Gribanov, N.~Y. Zolotykh, {Interpretable
  Feature Generation in ECG Using a Variational Autoencoder}, Frontiers in
  genetics 12 (2021) 638191.
\newblock \href {https://doi.org/https://doi.org/10.3389/fgene.2021.638191}
  {\path{doi:https://doi.org/10.3389/fgene.2021.638191}}.

\bibitem{gyawali2018learning}
P.~K. Gyawali, B.~M. Horacek, J.~L. Sapp, L.~Wang, {Learning disentangled
  representation from 12-lead electrograms: application in localizing the
  origin of Ventricular Tachycardia}, arXiv preprint arXiv:1808.01524 (2018).
\newblock \href {https://doi.org/https://doi.org/10.48550/arXiv.1808.01524}
  {\path{doi:https://doi.org/10.48550/arXiv.1808.01524}}.

\bibitem{lee2021conditional}
J.~Lee, W.~Kim, D.~Gwak, E.~Choi, Conditional generation of periodic signals
  with fourier-based decoder, arXiv preprint arXiv:2110.12365 (2021).
\newblock \href {https://doi.org/https://doi.org/10.48550/arXiv.2110.12365}
  {\path{doi:https://doi.org/10.48550/arXiv.2110.12365}}.

\bibitem{Robust_Anomaly_2021}
P.~Matias, D.~Folgado, H.~Gamboa, A.~Carreiro, Robust anomaly detection in time
  series through variational autoencoders and a local similarity score, in:
  Biosignals, 2021, pp. 91--102.
\newblock \href {https://doi.org/10.5220/0010320500910102}
  {\path{doi:10.5220/0010320500910102}}.

\bibitem{pereira2019unsupervised}
J.~Pereira, M.~Silveira, {Unsupervised representation learning and anomaly
  detection in ECG sequences}, International Journal of Data Mining and
  Bioinformatics 22~(4) (2019) 389--407.
\newblock \href {https://doi.org/https://doi.org/10.1504/IJDMB.2019.101395}
  {\path{doi:https://doi.org/10.1504/IJDMB.2019.101395}}.

\bibitem{raza2023anofed}
A.~Raza, K.~P. Tran, L.~Koehl, S.~Li, Anofed: Adaptive anomaly detection for
  digital health using transformer-based federated learning and support vector
  data description, Engineering Applications of Artificial Intelligence 121
  (2023) 106051.
\newblock \href
  {https://doi.org/https://doi.org/10.1016/j.engappai.2023.106051}
  {\path{doi:https://doi.org/10.1016/j.engappai.2023.106051}}.

\bibitem{zhu2021ecg}
J.~Zhu, W.~Fan, {ECG Data Modeling and Analyzing via Deep Representation
  Learning and Nonparametric Hidden Markov Models}, in: Proceedings of the 44th
  International ACM SIGIR Conference on Research and Development in Information
  Retrieval, 2021, pp. 1905--1909.
\newblock \href {https://doi.org/https://doi.org/10.1145/3404835.3463044}
  {\path{doi:https://doi.org/10.1145/3404835.3463044}}.

\bibitem{todo2023counterfactual}
W.~Todo, M.~Selmani, B.~Laurent, J.-M. Loubes, {Counterfactual Explanation for
  Multivariate Times Series Using A Contrastive Variational Autoencoder}, in:
  ICASSP 2023-2023 IEEE International Conference on Acoustics, Speech and
  Signal Processing (ICASSP), IEEE, 2023, pp. 1--5.
\newblock \href {https://doi.org/10.1109/ICASSP49357.2023.10095789}
  {\path{doi:10.1109/ICASSP49357.2023.10095789}}.

\bibitem{li2023descod}
H.~Li, G.~Ditzler, J.~Roveda, A.~Li, {DeScoD-ECG: Deep Score-Based Diffusion
  Model for ECG Baseline Wander and Noise Removal}, IEEE Journal of Biomedical
  and Health Informatics (2023).
\newblock \href {https://doi.org/10.1109/JBHI.2023.3237712}
  {\path{doi:10.1109/JBHI.2023.3237712}}.

\bibitem{alcaraz2023diffusion}
J.~M.~L. Alcaraz, N.~Strodthoff, {Diffusion-based Conditional ECG Generation
  with Structured State Space Models}, arXiv preprint arXiv:2301.08227 (2023).
\newblock \href
  {https://doi.org/https://doi.org/10.1016/j.compbiomed.2023.107115}
  {\path{doi:https://doi.org/10.1016/j.compbiomed.2023.107115}}.

\bibitem{adib2023synthetic}
E.~Adib, A.~Fernandez, F.~Afghah, J.~J. Prevost, {Synthetic ECG Signal
  Generation using Probabilistic Diffusion Models}, arXiv preprint
  arXiv:2303.02475 (2023).
\newblock \href {https://doi.org/https://doi.org/10.48550/arXiv.2303.02475}
  {\path{doi:https://doi.org/10.48550/arXiv.2303.02475}}.

\bibitem{alcaraz2022diffusion}
J.~M.~L. Alcaraz, N.~Strodthoff, {Diffusion-based Time Series Imputation and
  Forecasting with Structured State Space Models }, arXiv preprint
  arXiv:2208.09399 (2022).
\newblock \href {https://doi.org/https://doi.org/10.48550/arXiv.2208.09399}
  {\path{doi:https://doi.org/10.48550/arXiv.2208.09399}}.

\bibitem{neifar2024diffecg}
N.~Neifar, A.~Ben-Hamadou, A.~Mdhaffar, M.~Jmaiel, Diffecg: A versatile
  probabilistic diffusion model for ecg signals synthesis, in: 2024 IEEE/ACIS
  22nd International Conference on Software Engineering Research, Management
  and Applications (SERA), IEEE, 2024, pp. 182--188.
\newblock \href
  {https://doi.org/https://doi.org/10.1109/SERA61261.2024.10685651}
  {\path{doi:https://doi.org/10.1109/SERA61261.2024.10685651}}.

\bibitem{s23198328}
M.~H. Zama, F.~Schwenker, {ECG Synthesis via Diffusion-Based State Space
  Augmented Transformer}, Sensors 23~(19) (2023).
\newblock \href {https://doi.org/10.3390/s23198328}
  {\path{doi:10.3390/s23198328}}.

\bibitem{an2022auto}
Y.~An, H.~K. Lam, S.~H. Ling, {Auto-Denoising for EEG Signals Using Generative
  Adversarial Network}, Sensors 22~(5) (2022) 1750.
\newblock \href {https://doi.org/https://doi.org/10.3390/s22051750}
  {\path{doi:https://doi.org/10.3390/s22051750}}.

\bibitem{kwon2022novel}
J.~Kwon, C.-H. Im, {Novel Signal-to-Signal translation method based on StarGAN
  to generate artificial EEG for SSVEP-based brain-computer interfaces}, Expert
  Systems with Applications 203 (2022) 117574.
\newblock \href {https://doi.org/https://doi.org/10.1016/j.eswa.2022.117574}
  {\path{doi:https://doi.org/10.1016/j.eswa.2022.117574}}.

\bibitem{du2021multimodal}
C.~Du, C.~Du, H.~He, Multimodal deep generative adversarial models for scalable
  doubly semi-supervised learning, Information Fusion 68 (2021) 118--130.
\newblock \href {https://doi.org/https://doi.org/10.1016/j.inffus.2020.11.003}
  {\path{doi:https://doi.org/10.1016/j.inffus.2020.11.003}}.

\bibitem{calhas2020eeg}
D.~Calhas, R.~Henriques, {EEG to fMRI Synthesis: Is Deep Learning a
  candidate?}, arXiv preprint arXiv:2009.14133 (2020).
\newblock \href {https://doi.org/https://doi.org/10.48550/arXiv.2009.14133}
  {\path{doi:https://doi.org/10.48550/arXiv.2009.14133}}.

\bibitem{geng2021auxiliary}
D.~Geng, Z.~S. Chen, {Auxiliary Classifier Generative Adversarial Network for
  Interictal Epileptiform Discharge Modeling and EEG Data Augmentation}, in:
  2021 10th International IEEE/EMBS Conference on Neural Engineering (NER),
  IEEE, 2021, pp. 1130--1133.
\newblock \href {https://doi.org/10.1109/NER49283.2021.9441359}
  {\path{doi:10.1109/NER49283.2021.9441359}}.

\bibitem{lee2021contextual}
W.~Lee, J.~Lee, Y.~Kim, {Contextual imputation with missing sequence of EEG
  signals using generative adversarial networks}, IEEE Access 9 (2021)
  151753--151765.
\newblock \href {https://doi.org/10.1109/ACCESS.2021.3126345}
  {\path{doi:10.1109/ACCESS.2021.3126345}}.

\bibitem{zhang2021eeg}
A.~Zhang, L.~Su, Y.~Zhang, Y.~Fu, L.~Wu, S.~Liang, {EEG data augmentation for
  emotion recognition with a multiple generator conditional Wasserstein GAN},
  Complex \& Intelligent Systems (2021) 1--13\href
  {https://doi.org/https://doi.org/10.1007/s40747-021-00336-7}
  {\path{doi:https://doi.org/10.1007/s40747-021-00336-7}}.

\bibitem{liu2023eeg}
Q.~Liu, J.~Hao, Y.~Guo, {EEG Data Augmentation for Emotion Recognition with a
  Task-Driven GAN}, Algorithms 16~(2) (2023) 118.
\newblock \href {https://doi.org/https://doi.org/10.3390/a16020118}
  {\path{doi:https://doi.org/10.3390/a16020118}}.

\bibitem{kalashami2022eeg}
M.~P. Kalashami, M.~M. Pedram, H.~Sadr, {EEG Feature Extraction and Data
  Augmentation in Emotion Recognition}, Computational Intelligence and
  Neuroscience 2022 (2022).
\newblock \href {https://doi.org/https://doi.org/10.1155/2022/7028517}
  {\path{doi:https://doi.org/10.1155/2022/7028517}}.

\bibitem{hu2022e2sgan}
M.~Hu, J.~Chen, S.~Jiang, W.~Ji, S.~Mei, L.~Chen, X.~Wang, {E2SGAN: EEG-to-SEEG
  translation with generative adversarial networks}, Frontiers in Neuroscience
  16 (2022) 971829.
\newblock \href {https://doi.org/https://doi.org/10.3389/fnins.2022.971829}
  {\path{doi:https://doi.org/10.3389/fnins.2022.971829}}.

\bibitem{song2021improving}
Z.~Song, J.~Wang, G.~Yi, B.~Deng, {Improving EEG-based Alzheimer’s Disease
  Identification with Generative Adversarial Learning}, in: 2021 40th Chinese
  Control Conference (CCC), IEEE, 2021, pp. 3351--3356.
\newblock \href {https://doi.org/10.23919/CCC52363.2021.9550108}
  {\path{doi:10.23919/CCC52363.2021.9550108}}.

\bibitem{truong2019epileptic}
N.~D. Truong, L.~Kuhlmann, M.~R. Bonyadi, D.~Querlioz, L.~Zhou, O.~Kavehei,
  Epileptic seizure forecasting with generative adversarial networks, IEEE
  Access 7 (2019) 143999--144009.
\newblock \href {https://doi.org/10.1109/ACCESS.2019.2944691}
  {\path{doi:10.1109/ACCESS.2019.2944691}}.

\bibitem{xu2022bwgan}
M.~Xu, Y.~Chen, Y.~Wang, D.~Wang, Z.~Liu, L.~Zhang, {BWGAN-GP: An EEG data
  generation method for class imbalance problem in RSVP tasks}, IEEE
  Transactions on Neural Systems and Rehabilitation Engineering 30 (2022)
  251--263.
\newblock \href {https://doi.org/10.1109/TNSRE.2022.3145515}
  {\path{doi:10.1109/TNSRE.2022.3145515}}.

\bibitem{zhang2021realizing}
X.~Zhang, Z.~Lu, T.~Zhang, H.~Li, Y.~Wang, Q.~Tao, Realizing the application of
  eeg modeling in bci classification: based on a conditional gan converter,
  Frontiers in Neuroscience (2021) 1421\href
  {https://doi.org/https://doi.org/10.3389/fnins.2021.727394}
  {\path{doi:https://doi.org/10.3389/fnins.2021.727394}}.

\bibitem{biswas2023characterization}
S.~Biswas, P.~Chand, A.~Mathur, R.~Sinha, {Characterization of the
  event-related potentials during GAN-based generation of EEG signals and their
  data augmented subject classification}, in: 2023 International Conference on
  Recent Advances in Electrical, Electronics \& Digital Healthcare Technologies
  (REEDCON), IEEE, 2023, pp. 717--722.
\newblock \href {https://doi.org/10.1109/REEDCON57544.2023.10151321}
  {\path{doi:10.1109/REEDCON57544.2023.10151321}}.

\bibitem{prabowo2023advanced}
D.~W. Prabowo, H.~A. Nugroho, N.~A. Setiawan, J.~Debayle, {An Advanced Data
  Augmentation Scheme on Limited EEG Signals for Human Emotion Recognition},
  in: Proceeding of the 3rd International Conference on Electronics, Biomedical
  Engineering, and Health Informatics: ICEBEHI 2022, 5--6 October, Surabaya,
  Indonesia, Springer, 2023, pp. 391--409.
\newblock \href {https://doi.org/https://doi.org/10.1007/978-981-99-0248-4_27}
  {\path{doi:https://doi.org/10.1007/978-981-99-0248-4_27}}.

\bibitem{hwaidi2021noise}
J.~F. Hwaidi, T.~M. Chen, {A Noise Removal Approach from EEG Recordings Based
  on Variational Autoencoders}, in: 2021 13th International Conference on
  Computer and Automation Engineering (ICCAE), IEEE, 2021, pp. 19--23.
\newblock \href {https://doi.org/10.1109/ICCAE51876.2021.9426150}
  {\path{doi:10.1109/ICCAE51876.2021.9426150}}.

\bibitem{behrouzi2022graph}
T.~Behrouzi, D.~Hatzinakos, {Graph variational auto-encoder for deriving
  EEG-based graph embedding}, Pattern Recognition 121 (2022) 108202.
\newblock \href {https://doi.org/https://doi.org/10.1016/j.patcog.2021.108202}
  {\path{doi:https://doi.org/10.1016/j.patcog.2021.108202}}.

\bibitem{bethge2022eeg2vec}
D.~Bethge, P.~Hallgarten, T.~Grosse-Puppendahl, M.~Kari, L.~L. Chuang,
  O.~{\"O}zdenizci, A.~Schmidt, {EEG2Vec: Learning Affective EEG
  Representations via Variational Autoencoders}, in: 2022 IEEE International
  Conference on Systems, Man, and Cybernetics (SMC), IEEE, 2022, pp.
  3150--3157.
\newblock \href {https://doi.org/10.1109/SMC53654.2022.9945517}
  {\path{doi:10.1109/SMC53654.2022.9945517}}.

\bibitem{ahmed2022examining}
T.~Ahmed, L.~Longo, {Examining the Size of the Latent Space of Convolutional
  Variational Autoencoders Trained With Spectral Topographic Maps of EEG
  Frequency Bands}, IEEE Access 10 (2022) 107575--107586.
\newblock \href {https://doi.org/10.1109/ACCESS.2022.3212777}
  {\path{doi:10.1109/ACCESS.2022.3212777}}.

\bibitem{li2020latent}
X.~Li, Z.~Zhao, D.~Song, Y.~Zhang, J.~Pan, L.~Wu, J.~Huo, C.~Niu, D.~Wang,
  {Latent factor decoding of multi-channel EEG for emotion recognition through
  autoencoder-like neural networks}, Frontiers in neuroscience 14 (2020) 87.
\newblock \href {https://doi.org/https://doi.org/10.3389/fnins.2020.00087}
  {\path{doi:https://doi.org/10.3389/fnins.2020.00087}}.

\bibitem{tian2023dual}
C.~Tian, Y.~Ma, J.~Cammon, F.~Fang, Y.~Zhang, M.~Meng, {Dual-Encoder VAE-GAN
  With Spatiotemporal Features for Emotional EEG Data Augmentation}, IEEE
  Transactions on Neural Systems and Rehabilitation Engineering (2023).
\newblock \href {https://doi.org/10.1109/TNSRE.2023.3266810}
  {\path{doi:10.1109/TNSRE.2023.3266810}}.

\bibitem{tosato2023eeg}
G.~Tosato, C.~M. Dalbagno, F.~Fumagalli, {EEG Synthetic Data Generation Using
  Probabilistic Diffusion Models}, arXiv preprint arXiv:2303.06068 (2023).
\newblock \href {https://doi.org/https://doi.org/10.48550/arXiv.2303.06068}
  {\path{doi:https://doi.org/10.48550/arXiv.2303.06068}}.

\bibitem{AnAccurateNonAccelerometer}
A.~H. Afandizadeh, S.~A.~H. Aqajari, H.~Khodabandeh, A.~Rahmani, F.~Kurdahi,
  {An Accurate Non-Accelerometer-Based PPG Motion Artifact Removal Technique
  Using CycleGAN}, ACM Trans. Comput. Healthcare 4~(1) (2023).
\newblock \href {https://doi.org/https://doi.org/10.1145/3563949}
  {\path{doi:https://doi.org/10.1145/3563949}}.

\bibitem{vo2021p2e}
K.~Vo, E.~K. Naeini, A.~Naderi, D.~Jilani, A.~M. Rahmani, N.~Dutt, H.~Cao,
  {P2E-WGAN: ECG waveform synthesis from PPG with conditional wasserstein
  generative adversarial networks}, in: Proceedings of the 36th Annual ACM
  Symposium on Applied Computing, 2021, pp. 1030--1036.
\newblock \href {https://doi.org/https://doi.org/10.1145/3412841.3441979}
  {\path{doi:https://doi.org/10.1145/3412841.3441979}}.

\bibitem{hwang2022user}
D.~Y. Hwang, {User Recognition System Based on PPG Signal}, Ph.D. thesis,
  University of Toronto (Canada) (2022).

\bibitem{sarkar2021cardiogan}
P.~Sarkar, A.~Etemad, {CardioGAN: Attentive Generative Adversarial Network with
  Dual Discriminators for Synthesis of ECG from PPG}, in: Proceedings of the
  AAAI Conference on Artificial Intelligence, Vol.~35, 2021, pp. 488--496.
\newblock \href {https://doi.org/https://doi.org/10.1609/aaai.v35i1.16126}
  {\path{doi:https://doi.org/10.1609/aaai.v35i1.16126}}.

\bibitem{mendez2022emg}
V.~Mendez, C.~Lhoste, S.~Micera, {EMG Data Augmentation for Grasp
  Classification Using Generative Adversarial Networks}, in: 2022 44th Annual
  International Conference of the IEEE Engineering in Medicine \& Biology
  Society (EMBC), IEEE, 2022, pp. 3619--3622.
\newblock \href {https://doi.org/10.1109/EMBC48229.2022.9871625}
  {\path{doi:10.1109/EMBC48229.2022.9871625}}.

\bibitem{olsson2021can}
A.~E. Olsson, N.~Male{\v{s}}evi{\'c}, A.~Bj{\"o}rkman, C.~Antfolk, {Can Deep
  Synthesis of EMG Overcome the Geometric Growth of Training Data Required to
  Recognize Multiarticulate Motions?}, in: 2021 43rd Annual International
  Conference of the IEEE Engineering in Medicine \& Biology Society (EMBC),
  IEEE, 2021, pp. 6380--6383.
\newblock \href {https://doi.org/10.1109/EMBC46164.2021.9630276}
  {\path{doi:10.1109/EMBC46164.2021.9630276}}.

\bibitem{goodfellow2020generative}
I.~Goodfellow, J.~Pouget-Abadie, M.~Mirza, B.~Xu, D.~Warde-Farley, S.~Ozair,
  A.~Courville, Y.~Bengio, Generative adversarial networks, Communications of
  the ACM 63~(11) (2020) 139--144.
\newblock \href {https://doi.org/https://doi.org/10.1145/3422622}
  {\path{doi:https://doi.org/10.1145/3422622}}.

\bibitem{kingma2013auto}
D.~P. Kingma, M.~Welling, {Auto-Encoding Variational Bayes}, arXiv preprint
  arXiv:1312.6114 (2013).
\newblock \href {https://doi.org/https://doi.org/10.48550/arXiv.1312.6114}
  {\path{doi:https://doi.org/10.48550/arXiv.1312.6114}}.

\bibitem{ho2020denoising}
J.~Ho, A.~Jain, P.~Abbeel, {Denoising Diffusion Probabilistic Models}, Advances
  in Neural Information Processing Systems 33 (2020) 6840--6851.

\bibitem{song2019generative}
Y.~Song, S.~Ermon, {Generative Modeling by Estimating Gradients of the Data
  Distribution}, Advances in neural information processing systems 32 (2019).

\bibitem{khader2023denoising}
F.~Khader, G.~M{\"u}ller-Franzes, S.~Tayebi~Arasteh, T.~Han, C.~Haarburger,
  M.~Schulze-Hagen, P.~Schad, S.~Engelhardt, B.~Bae{\ss}ler, S.~Foersch,
  et~al., {Denoising diffusion probabilistic models for 3D medical image
  generation}, Scientific Reports 13~(1) (2023) 7303.

\bibitem{mitbih}
G.~Moody, R.~Mark, The impact of the mit-bih arrhythmia database, IEEE
  Engineering in Medicine and Biology Magazine 20~(3) (2001) 45--50.
\newblock \href {https://doi.org/10.1109/51.932724}
  {\path{doi:10.1109/51.932724}}.

\bibitem{moody1984noise}
G.~B. Moody, W.~Muldrow, R.~G. Mark, A noise stress test for arrhythmia
  detectors, Computers in cardiology 11~(3) (1984) 381--384.

\bibitem{wagner2020ptb}
P.~Wagner, N.~Strodthoff, R.-D. Bousseljot, D.~Kreiseler, F.~I. Lunze,
  W.~Samek, T.~Schaeffter, {PTB-XL, a large publicly available
  electrocardiography dataset}, Scientific data 7~(1) (2020) 154.
\newblock \href {https://doi.org/https://doi.org/10.1038/s41597-020-0495-6}
  {\path{doi:https://doi.org/10.1038/s41597-020-0495-6}}.

\bibitem{BIDMC}
M.~A.~F. Pimentel, A.~E.~W. Johnson, P.~H. Charlton, D.~Birrenkott, P.~J.
  Watkinson, L.~Tarassenko, D.~A. Clifton, {Toward a Robust Estimation of
  Respiratory Rate From Pulse Oximeters}, IEEE Transactions on Biomedical
  Engineering 64~(8) (2017) 1914--1923.
\newblock \href {https://doi.org/10.1109/TBME.2016.2613124}
  {\path{doi:10.1109/TBME.2016.2613124}}.

\end{thebibliography}
\end{document}